\documentclass[11pt]{article}

\usepackage[preprint]{acl}

\usepackage{times}
\usepackage{latexsym}

\usepackage[T1]{fontenc}

\usepackage[utf8]{inputenc}

\usepackage{microtype}

\usepackage{inconsolata}

\usepackage{graphicx}
\usepackage{amsmath}
\usepackage{booktabs}
\usepackage{pifont}
\usepackage{tikz}
\usepackage[table]{xcolor}
\usepackage[skins,breakable]{tcolorbox}
\usepackage{soul}
\usepackage{array}
\usetikzlibrary{positioning, fit, shapes.geometric, calc, backgrounds}

\definecolor{diffremoved}{RGB}{255,200,200}
\definecolor{diffadded}{RGB}{200,255,200}
\definecolor{textgray}{RGB}{50,50,50}
\definecolor{bordercolor}{RGB}{180,180,180}

\newcommand{\removed}[1]{{\sethlcolor{diffremoved}\hl{#1}}}
\newcommand{\added}[1]{{\sethlcolor{diffadded}\hl{#1}}}

\newtcolorbox{samplebox}{
  enhanced,
  colback=white,
  boxrule=0.75pt,
  arc=2pt,
  left=4pt,
  right=4pt,
  top=4pt,
  bottom=4pt,
  fontupper=\small
}

\title{Reasoning Gets Harder for LLMs Inside A Dialogue}

\author{Ivan Kartáč {\normalfont \and}  Mateusz Lango {\normalfont \and} Ondřej Dušek \\[2pt]
Charles University, Faculty of Mathematics and Physics \\ Institute of Formal and Applied Linguistics \\
Prague, Czechia \\[2pt]
\texttt{\{kartac,lango,odusek\}@ufal.mff.cuni.cz}
}

\begin{document}
\maketitle
\begin{abstract}
Large Language Models (LLMs) achieve strong performance on many reasoning benchmarks, yet these evaluations typically focus on isolated tasks that differ from real-world usage in task-oriented dialogue (TOD). In this setting, LLMs must perform reasoning inherently while generating text and adhering to instructions on role, format, and style. This mismatch raises concerns about whether benchmark performance accurately reflects models’ reasoning robustness in TOD setting. We investigate how framing reasoning tasks within TOD affects LLM performance by introducing \textsc{Boulder}, a new dynamic benchmark covering eight travel-related tasks that require arithmetic, spatial, and temporal reasoning with both commonsense and formal aspects. Each problem is presented in both isolated and dialogue-based variants, enabling controlled comparison while mitigating data contamination. Experiments on eight LLMs reveal a substantial and consistent performance gap between isolated and dialogue settings. Through ablations and qualitative analysis, we show that this gap is largely driven by the multi-turn nature of dialogue, with additional effects from role conditioning and tool-use requirements. Our results highlight the need to evaluate LLM reasoning in realistic interactive scenarios.\footnote{The benchmark, code, and model outputs are available at \url{https://github.com/ivankartac/boulder}}
\end{abstract}

\section{Introduction}

\begin{figure}[t!]
\centering
\includegraphics[width=\columnwidth]{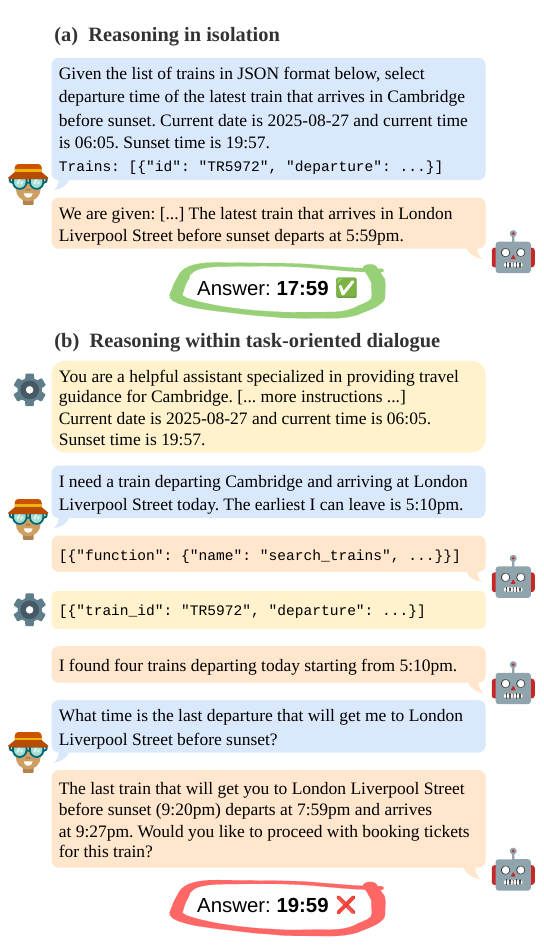}
\caption{An example from our \textsc{Boulder} benchmark, showing the same problem instance in two variants: as an isolated task and within a task-oriented dialogue.}
\label{fig:evaluation-examples}
\end{figure}

Large Language Models (LLMs) have recently demonstrated promising results on a number of complex reasoning benchmarks (\citealp{deepseekai2025deepseekr1incentivizingreasoningcapability,yang2025qwen3technicalreport}). However, these benchmarks usually evaluate LLMs on well-defined isolated tasks (\citealp{hendrycks2021measuring,rein2024gpqa,chollet2025arc}), a setup which often differs from their practical real-world use. In contrast, evaluating reasoning capabilities of LLMs as part of a broader task has not yet been sufficiently explored (\citealp{cui-etal-2020-mutual,li2023diplomat}). Within language processing tasks, LLMs are typically expected to adhere to various instructions regarding their role, output format, style, or response length. At the same time, the model must implicitly perform reasoning to provide reliable responses: for example, machine translation may require localization depending on the source and target language (e.g., conversion of values to different units), which requires reasoning to successfully complete the task.
As a result, strong performance on traditional benchmarks may overestimate LLMs robustness in realistic interactive scenarios.

The main goals of this work are to investigate how framing a reasoning task within a task-oriented dialogue (TOD) setting affects the reasoning capabilities of LLMs, and to explain where any potential performance gaps stem from.
We focus on TOD because it is a challenging task that combines various aspects of text generation within a single framework.
First, TOD is inherently multi-turn, allowing us to investigate the effects of iterative task solving and generation. Second, it enables testing various reasoning capabilities, as TOD often requires commonsense reasoning, spatiotemporal reasoning, and reasoning over structured data extracted from databases. Third, TOD response generation is frequently constrained by length, style, or format, incorporating text generation guidelines that are not directly related to reasoning but are common in typical LLM usage.
Finally, TOD closely aligns our benchmark with real-world scenarios, as recent systems often employ a single LLM instead of a traditional modular pipeline, performing both tool calls and response generation in a zero-shot setting~\cite{xu-etal-2024-rethinking,li-etal-2024-large-language-models,baidya-etal-2025-behavior}.

We construct \textsc{Boulder} (\textbf{B}enchmarking \textbf{o}f \textbf{U}sefulness of \textbf{L}LMs in \textbf{D}ialogue-\textbf{E}mbedded \textbf{R}easoning), a benchmark with eight distinct tasks based on four travel-related domains, covering arithmetic, spatial, and temporal reasoning, with both common-sense and formal aspects. Each task is realistic and reflects the types of questions that a user might reasonably ask in an interaction with a task-oriented dialogue system. 
All examples are presented with the same reasoning problem instance in both isolated and a dialogue setting, making only minimal necessary differences in the assignment. 
To avoid LLM training data contamination (\citealp{sainz-etal-2023-nlp,balloccu-etal-2024-leak,han2025searchtime}), we design the benchmark as dynamic: examples can be freshly re-generated, while targets remain automatically verifiable.

In our experiments on the benchmark, we show that modern LLMs struggle to maintain the performance of the isolated setting in the dialogue scenarios. We find that the performance penalty can be explained by the multi-turn nature of TOD tasks, the role assumed by the LLM, and the dual function of the LLM where it is expected both to generate responses and perform tool calling.

To summarize, our contributions are the following:
\begin{itemize}
    \item We construct a new dynamic benchmark to test the reasoning capabilities of LLMs in realistic dialogue scenarios (see Section~\ref{sec:methods}). The benchmark is designed to test systems in both dialogue and isolated settings.
    \item We evaluate eight open-weight and proprietary LLMs of different sizes and architectures, and show a significant performance gap between the isolated and dialogue settings (Section~\ref{sec:experiments},~\ref{sec:results}).

    \item To identify failure modes and explain the observed discrepancy, we design a series of ablation experiments (Section~\ref{sec:ablations}) and conduct a qualitative analysis of model responses (Section~\ref{sec:results}).
\end{itemize}

\section{Related Work}
\label{sec:related}

\paragraph{LLMs in Task-oriented Dialogue} Although task-oriented dialogue systems have traditionally been implemented as modular pipelines, recent works have proposed the use of LLMs for this task \citep{hudecek-dusek-2023-large,stricker-paroubek-2024-shot,steindl-etal-2025-improved,joshi_controllable_2025,xu_sqlwoz_2025}, often replacing individual modules by a single LLM with tool calling (\citealp{li-etal-2024-large-language-models,xu-etal-2024-rethinking,acikgoz_can_2025}).
However, when evaluating their systems, these works typically focus on general conversational abilities and do not address reasoning specifically.

\paragraph{Reasoning in Dialogue}
Outside task-oriented dialogue, \citet{cui-etal-2020-mutual} presented a benchmark for reasoning about the speaker's attitude or intention, facts, or situations in multi-turn dialogue, measured by next-utterance selection from four candidates. 
CoQA \citep{reddy-etal-2019-coqa} frames reading comprehension as conversational question answering that includes pragmatic reasoning. 
Diplomat \citep{li2023diplomat} is another dataset for evaluation of pragmatic reasoning in dialogue, while TimeDial \citep{qin-etal-2021-timedial} focuses on temporal common-sense reasoning. All of these datasets use multiple-choice answers as a proxy for reasoning evaluation. While such a setup makes it easy to verify model answers, it is not clear to what extent the results on such benchmarks translate to real-world performance of evaluated models \citep{liao2021are}. In contrast, our work evaluates reasoning embedded in language generation, which is a more realistic setting.

\paragraph{Complex instructions and multi-turn interaction}
Several recent benchmarks focus on evaluating the ability of LLMs to follow instructions \citep{zhou2023instructionfollowingevaluationlargelanguage}, especially complex/compound ones \citep{he2024can,son-etal-2024-multi-task}. The evaluation often focuses on easily verifiable formal aspects (e.g., maintaining required text length).
\citet{coca-etal-2025-aspera} have a similar aim, but specifically focus on tasks solved by generating and executing code.
Others make tasks more challenging by adding in external factors, such as personality specification \cite{gupta2024bias} or asking for confirmation \cite{laban2024surechallengingllmsleads}, finding that this typically leads to performance drops.

There has been a slew of works that specifically focus on evaluating multi-turn interaction with LLMs \cite{bai-etal-2024-mt,deshpande-etal-2025-multichallenge,kwan-etal-2024-mt,wang2024mint}.
Their general finding is that multi-turn interaction reduces the ability of LLMs to follow instructions.
Unlike ours, these works do not focus on embedded tasks. 

\paragraph{Temporal reasoning}
There are several recent benchmarks for temporal reasoning evaluation.
The TRAM benchmark~\citep{wang-zhao-2024-tram} integrates 10 temporal datasets evaluating models' capabilities using multiple-choice questions. TimeBench \citep{chu-etal-2024-timebench} also involves several tasks, but only one of them, TimeDial \citep{qin-etal-2021-timedial}, considers dialogue data. However, as mentioned earlier, TimeDial also relies on selecting multiple-choice answers rather than on generating responses. \citet{islakoglu-kalo-2025-chronosense} leverage Allen’s interval relations to construct a benchmark of binary questions. More recently, Test-of-Time~\citep{fatemi2025test} employed synthetic, time-related questions, but dialogue data was not considered.

\paragraph{Spatial reasoning}
Spatial reasoning is often evaluated using multimodal data~\cite{7410636,hudson2019gqanewdatasetrealworld}, and research on spatial reasoning in text-only models is relatively limited.
\citet{mirzaee-etal-2021-spartqa} constructed simple descriptions of special situations and evaluated the capabilities of several BERT-based models.
\citet{patel2022mapping} measured whether language models preserve spatial capabilities under isomorphic transformations of the space.
\cite{momennejad2023evaluating,yamada2024evaluating} evaluated LLM capabilities on reasoning over grid maps with different connectivity patterns, but their work did not involve dialogue.

\section{The \textsc{Boulder} Benchmark}
\label{sec:methods}

We design our benchmark as dynamic and with automatically verifiable answers. Each example is presented in an isolated and a dialogue setting. In the dialogue setting, it consists of a fixed message history with previous user and assistant messages, and tool calls with retrieved results. The last user message corresponds to the main query which the LLM is expected to answer.
The baseline with an isolated task consists of a problem description (e.g. \emph{Given the data, what is the latest train that arrives in Cambridge before 6:15pm?}) and JSON data identical to those retrieved by tool calls in the dialogue setting. The question is formulated in the closest possible way to the one in the dialogue.

\begin{table*}[ht]
\centering
\small
\renewcommand{\arraystretch}{1.25}
\setlength{\tabcolsep}{4pt}
\begin{tabular}{lp{1.7cm}p{6.3cm}ll}
\toprule
\textbf{Task} & \textbf{Domains} & \textbf{Example query} & \textbf{Extracted value} & \textbf{Metric} \\
\midrule
Ticket price & \texttt{trains} & \emph{Can you first calculate the total price for me? There are four of us. Three of us are buying return tickets, one standard class, and two first class - one with a 33\% discount, the other without discount. Another person is buying a one-way ticket.} & amount (float) & Accuracy \\
Booking price & \texttt{hotels} & \emph{Can you first calculate the total price for me? It's for five people staying for four nights. One person will stay in a single room and the others will share double rooms. One person in a double room will check in one day later.} & amount (float) & Accuracy \\
\noalign{\vspace{0.25em}}
\hline
\noalign{\vspace{0.25em}}
Departure time & \texttt{trains} & \emph{What time is the last departure that will get me to Cambridge before sunset?} & time (HH:MM) & Accuracy \\
Departure frequency & \texttt{trains} & \emph{How often on average do trains run between 19:00 and midnight?} & minutes (float) & MAE \\
Opening hours & \texttt{restaurants} & \emph{Which of them are open on Wednesday for the entire time between 2pm and 6:30pm?} & venues (array[string]) & Precision \\
\noalign{\vspace{0.25em}}
\hline
\noalign{\vspace{0.25em}}
Distance & \texttt{hotels, restaurants} & \emph{How far is it from the hotel to the restaurant?} & distance (float) & MAE \\
Directional relations & \texttt{attractions, restaurants} & \emph{Is Museum of Archaelogy and Anthropology south of Pizza Express?} & yes/no (bool) & Accuracy \\
Shortest path & \texttt{attractions, hotels} & \emph{I want to visit Broughton House Gallery, King's College, and Camboats. I'll walk between them starting from the hotel and taking a taxi back from the last one. What order should I visit all of them in to minimize my walking distance?} & path (array[string]) & Accuracy \\
\bottomrule
\end{tabular}
\caption{Overview of tasks in the \textsc{Boulder} benchmark, including corresponding domains, representative examples of user queries, types of values extracted from the generated answers, and primary evaluation metrics.}
\label{tab:tasks}
\end{table*}

\subsection{Tasks}
\label{sec:tasks}

\textsc{Boulder} contains eight tasks covering arithmetic, spatial, and temporal reasoning, with both formal and common-sense aspects. For temporal tasks, reasoning about temporal relations, order, and frequency is tested. Spatial tasks include reasoning about directional relations, distances, and paths, with the map represented as a simple 2D plane without streets or other barriers.

We do not provide the LLMs any tools to directly help them solve the given reasoning problem. Similarly, none of the provided tools return answers (including partial ones) to any of these tasks.

Examples of queries, extracted values, and metrics for each task are presented in Table \ref{tab:tasks}. In the following, we provide a brief overview of these tasks, while Appendix~\ref{sec:appendix_tasks} describes all of them in detail.

\paragraph{Train ticket price} This task tests arithmetic reasoning with a strong common-sense component: the user asks for a total price calculation for a group of tickets for multiple people with different parameter choices: one-way/return ticket, 1st/2nd class, and various discounts. As the example provided in Table \ref{tab:tasks} indicates, the assistant is conceptually required to parse an arithmetic expression from a complex natural language utterance.

\paragraph{Hotel booking price} 
This task is similar to the previous one, but involves a different domain: hotel booking.
The assistant is asked about the total price given the number of people, number of nights, specific assignments to different room types (single/double/family) and potential exceptions, such as one of the guests leaving a day earlier.

\paragraph{Train departure time} To test reasoning about temporal order, we include a task that requires identification of the last train that arrives at its destination before a specific time. We use sunset time as a threshold, which is explicitly provided to the assistant in the prompt.

\paragraph{Train departure frequency} Reasoning about temporal frequencies is represented by this task, in which the LLM is asked about the average time interval between departures of trains on a specific route. To make the task reasonably challenging, we synthesize additional trains and different frequencies for different times of day to create irregularities in departure times.

\paragraph{Restaurant opening hours} This task tests comparison of a reference time interval (user query) with multiple intervals (opening hours of retrieved restaurants) that have different relations to the reference. The relations are based on the Allen's interval algebra \citep{allen1983maintaining}, a formalism for temporal reasoning that defines possible relations between time intervals.

\paragraph{Distance between venues} This task tests the calculation or estimation of distances between two points. After the assistant finds and recommends a hotel and a restaurant, the user asks about the distance between the restaurant and the hotel.

\paragraph{Directional relations between venues} To test reasoning about directional relations, the assistant is instructed to determine whether an attraction is located in a specific cardinal direction (north, south, east, or west) from a restaurant, where the direction in the query is selected randomly.

\paragraph{Shortest walking path} To represent reasoning about paths and spatial order, the benchmark includes a variant of the Travelling Salesman Problem~\cite{lawler1985traveling}. The assistant is asked to find and book a hotel, then look up information on two to four different attractions, and finally find the shortest walking path to visit all of them from the hotel, assuming the user returns by taxi.

\subsection{Benchmark Construction}
\label{sec:benchmark}

Each task is implemented as a procedure that allows for random generation of new examples based on a database and a conversation template. We use database from the MultiWOZ dataset \citep{budzianowski-etal-2018-multiwoz} slightly adapted to achieve greater diversity by synthesizing new data for certain tasks. See Appendix~\ref{sec:benchmark_construction_details} for details.

We use conversation templates \cite{reddy-etal-2019-coqa,laban2024surechallengingllmsleads,deng-etal-2024-multi} to generate dialogue contexts for each task: the template contains the history of previous turns, including tool calls and their results (see Figure \ref{fig:conversation_template} for an example). Each message in the template may include placeholders to be filled with actual values during test example generation. To increase diversity of these contexts, we design a base dialogue history template for each task and use an LLM to provide multiple paraphrases of the template: the LLM is instructed to preserve the meaning of each message and all placeholders. We validate meaning preservation manually and use regex validation for placeholders (see details in Appendix \ref{sec:appendix_paraphrases}).

To enable evaluation at scale, our test examples are synthesized with verifiable answers. While the answers are unambiguous, the LLM may still phrase them in a large number of ways. Since our goal is to evaluate LLMs in real-world settings, we cannot instruct models to generate their responses in a specific automatically parseable format, such as highlighting the answer with the commonly used \texttt{\textbackslash boxed}\{\} command.  Therefore, the answers are parsed from generated responses by specialized LLM-based parsers, described in Section \ref{sec:parser}.

\subsection{Answer Extraction}
\label{sec:parser}

To extract verifiable answers from LLM responses, we design a set of parsers based on prompting the Qwen~3 30B MoE LLM \cite{yang2025qwen3technicalreport}. Each parser is instructed to extract values of a specific semantic category from responses of evaluated LLMs and return them as JSON. If the answer is expected to consist of a closed set of categorical values, we provide the list of possible values to the parser. For example, in the shortest walking path task, we provide the list of all relevant venues to the parser and ask it to extract them in the order given in the response. See Appendix~\ref{sec:appendix_parser} for details on the parsers.

To validate the reliability of our parsers, we conducted a human evaluation in which we annotated the correctness of the extracted values. For each task, we manually evaluated 30 random examples for each of the three main setups (see Section~\ref{sec:evaluation_settings}) and evaluated LLMs (see Section~\ref{sec:models}), i.e., 720 examples for each task and 5,760 examples overall.
As the results in Table~\ref{tab:parser_evaluation} show, the parsers achieve accuracy between 95-99\% depending on task (see Figure~\ref{fig:results_parser} in Appendix~\ref{sec:appendix_parser_validation} for detailed results). The accuracy range is similar across different LLMs, with the exception of Gemini~2.5 Flash, where the score is slightly lower. We also measured Cohen's~$\kappa$ coefficient for inter-annotator agreement on a subset of 160 parsed values with double annotation. We obtained $\kappa = 0.94$ on the evaluation of the correctness of the parser.

Using an LLM-based parser makes the evaluation considerably more reliable compared to LLM-as-a-judge (\citealp{li-etal-2025-generation,zheng2023judging,kocmi-federmann-2023-large,kim2024prometheus}) while not relying on expensive and time-consuming human evaluation. However, since our parser still occasionally extracts incorrect values, this procedure might lead to biased estimates. To correct for bias, we apply \emph{prediction-powered inference} (PPI, \citealp{angelopoulos2023predictionpoweredinference}), which provides unbiased point estimates and more reliable confidence intervals when measurement errors are present. See Appendix~\ref{sec:appendix_bias_correction} for details.

\begin{table}[t]
\centering
\small
\setlength{\tabcolsep}{5pt}
\begin{tabular}{lc|lc}
\toprule
\textbf{Task} & \textbf{Acc.} & \textbf{Model} & \textbf{Acc.} \\
\midrule
Ticket price & 96.39\% & Qwen3 4B & 96.52\% \\
Booking price & 98.75\% & Mistral Small 24B & 98.47\% \\
Departure time & 98.61\% & Qwen3 30B & 98.05\% \\
Departure freq. & 97.50\% & Command A 111B & 98.61\% \\
Opening hours & 94.86\% & Qwen3 235B & 95.83\% \\
Distance & 96.11\% & DeepSeek V3.2 & 95.97\% \\
Directional rel. & 96.54\% & Gemini 2.5 Flash & 93.47\% \\
Shortest path & 98.06\% & Claude Sonnet 4.5 & 95.97\% \\
\bottomrule
\end{tabular}
\caption{Results of human evaluation of parsers aggregated by task and model. For each task, we sampled 90 examples for each model, one for each of the three main setups.}
\label{tab:parser_evaluation}
\end{table}

\begin{table}[t]
\centering
\small
\begin{tabular}{lcc}
\toprule
\textbf{Model} & \textbf{Architecture} & \textbf{Open-weight} \\
\midrule
Qwen3 4B & Dense & \ding{51} \\
Mistral Small 24B & Dense & \ding{51} \\
Qwen3 30B A3B & MoE & \ding{51} \\
Command A 111B & Dense & \ding{51} \\
Qwen3 235B A22B & MoE & \ding{51} \\
DeepSeek V3.2 671B & MoE & \ding{51} \\
\hline
\noalign{\vspace{0.25em}}
Gemini 2.5 Flash & - & \ding{55} \\
Claude 4.5 Sonnet & - & \ding{55} \\
\bottomrule
\end{tabular}
\caption{Overview of the evaluated LLMs.}
\label{tab:models}
\end{table}

\section{Experiments}
\label{sec:experiments}

\subsection{Evaluation Settings}
\label{sec:evaluation_settings}

We use a standalone and a dialogue setting in our main experiments, both of which receive the same context presented either as an isolated prompt, or embedded in a TOD prompt and message history.
We frame the embedding in a dialogue task following \citet{xu-etal-2024-rethinking}, who use a single end-to-end response generation LLM prompt with function calling. This setup allows us to focus on reasoning capabilities of LLMs without the need to implement a complex modular dialogue pipeline. Unlike \citet{xu-etal-2024-rethinking}, we provide tools as JSON schema appended to the prompt. Also, we use standard prompting instead of ReAct prompting in our main experiments. The prompt templates and tools used for the TOD system are presented in Appendix~\ref{sec:appendix_dialogue_prompt}.
Our main experiments involve three different evaluation settings:

\begin{enumerate}
    \item \textbf{Baseline:} This is the standalone setup where the LLMs are prompted to solve the task in isolation, without any embedding in dialogue.
    
    \item \textbf{Dialogue:} This is a basic version that involves the dialogue framing.
    In this setup, we only instruct the model about the specific tasks and domain, without any response format, style or length instructions.

    \item \textbf{Dialogue-concise:} This setup also uses the dialogue framing, but LLMs are additionally instructed to provide concise responses and generate at most two sentences. This is the strictest setting, serving for additional comparison -- as the instructions for \emph{Dialogue} do not contain any length restrictions, the models are generally expected to score better there than on \emph{Dialogue-concise}. However, note that this kind of setup is often necessary for real-time dialogue, especially in spoken interaction.
\end{enumerate}

We generate 100 examples for each task described in Section~\ref{sec:tasks}, obtaining a total of 800 test examples.

Ultimately, the goal of our work is to understand the effect of specific aspects of the dialogue setup on the LLM reasoning performance. Therefore, we additionally conduct a series of ablation experiments, where we gradually change different attributes of the task, such as multi-turn/single-turn generation, presence of tools, or instructions related to the assistant's role. The ablations and their results are presented in detail in Section~\ref{sec:ablations}.

\subsection{Evaluated Models}
\label{sec:models}

We evaluate eight LLMs of different sizes and architectures. The models with their parameter sizes, architectures, and licenses are listed in Table~\ref{tab:models}.

We run inference for all open-weight models (except DeepSeek~V3.2) through the Ollama platform.\footnote{\url{https://ollama.com/}} To make the inference efficient, we run quantized versions of some of these models.\footnote{Note that while quantization can have an effect on absolute scores in our experiments, we primarily compare each model against itself in different settings, and our findings are consistent across both quantized and non-quantized models.} Details about specific versions and quantizations are provided in Appendix~\ref{sec:appendix_models}.
For both proprietary LLMs, we run the inference through the OpenRouter API.\footnote{\url{https://openrouter.ai/}} We also use OpenRouter for DeepSeek V3.2, as this model's 671B parameters are beyond our hardware's capabilities.
To ensure reproducibility, we use greedy decoding for all models in all settings.
For all ablation experiments, we apply the same setup as for the main evaluations.

\subsection{Metrics}

For the \emph{Train ticket price} and \emph{Hotel booking price} tasks which involve numerical quantities, we use \textbf{accuracy} computed by exact match (with rounding to integers, which only affects ticket prices in practice). The answers in the \emph{Train departure time} task are timestamps, also compared by exact match. Exact match is also applied to the \emph{Directional relations} task, where the answers are categorical (yes/no/unknown), and to the \emph{Shortest walking path} task, where we compare the order of venues with the target order. To evaluate the \emph{Restaurant opening hours} task, we compute the \textbf{precision} of listed restaurants with respect to targets. Finally, we use \textbf{mean absolute error (MAE)} as the metric for \emph{Distance between venues} and \emph{Train departure frequency}, measuring the error between distances in meters and frequencies in minutes, respectively. 

When aggregating results across tasks with different metrics, we use raw values for exact match and precision, and we normalize and invert MAE to a range between 0 and~1, where 1 corresponds to no error. We use the following formula for the normalization:

\[
\text{E}_{\text{inv+norm}}(y, \hat{y}) = \frac{1}{1 + \frac{\text{E}(y, \hat{y})}{k}}
\]

\noindent where $E(y, \hat{y})$ is an error between the answer and target value, and $k$ is set to 5 for minutes and 500 for meters. The value of $k$ can be interpreted as a marginally acceptable error, and MAE equal to this value will be transformed to a score of 0.5.\footnote{There is no effect on the statistical significance of the difference between the \emph{Baseline} and \emph{Dialogue} setups in the aggregate results shown in Figure~\ref{fig:results_main} even with values of $k$ orders of magnitude higher than this.}

For outputs in tasks evaluated with MAE where the models did not provide any answer, we set the parsed value to $\infty$, which is then transformed to 0 in the normalized space.

\section{Results}
\label{sec:results}

\begin{figure}[t!]
    \centering
\includegraphics[width=\columnwidth]{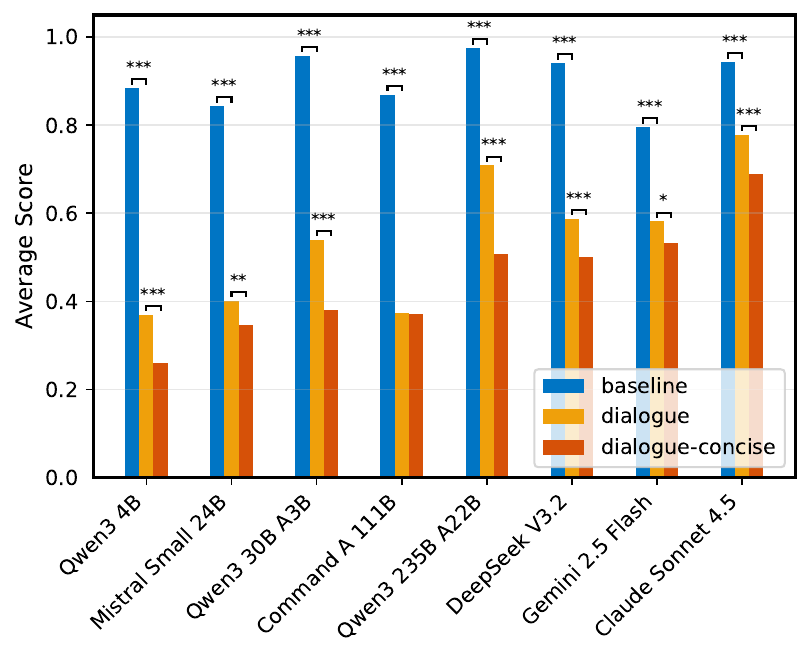}
    \caption{Evaluation results of 8 LLMs in the three main settings, averaged over all tasks. Asterisks indicate statistically significant differences between the \textit{Baseline} and \textit{Dialogue}, and \textit{Dialogue} and \textit{Dialogue-concise} settings (t-test, $*\colon p < 0.05$, $**\colon p < 0.01$, ${*}{*}{*}\colon p < 0.001$).}
    \label{fig:results_main}
\end{figure}

\begin{figure}[t]
    \centering
    \includegraphics[width=\columnwidth]{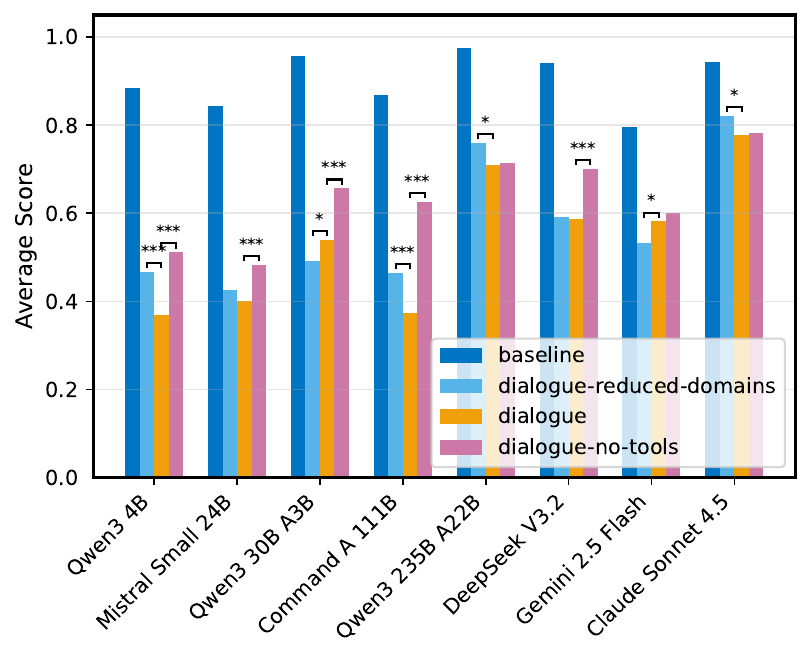}
    \caption{Results for the \textit{Dialogue with reduced domains} and \textit{Dialogue without tools} ablations, micro-averaged over all tasks. Significance indication follows Figure~\ref{fig:results_main}, and each ablation is compared to the main \emph{Dialogue} setup.}
    \label{fig:results_ablation_tools}
\end{figure}

\begin{figure*}[t]
    \centering
    \includegraphics[width=\textwidth]{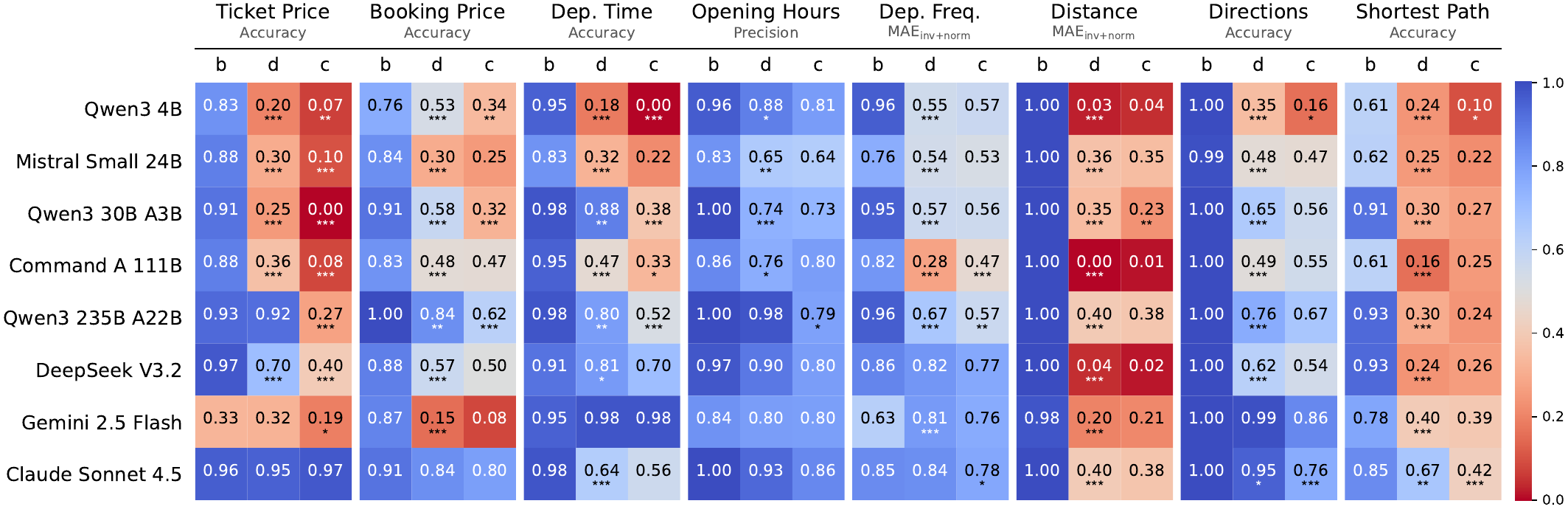}
    \caption{Detailed per-task results for the \emph{Baseline} (b), \emph{Dialogue} (d), and \emph{Dialogue-concise} (c) setups. Asterisks indicate significant differences between setups in neighboring columns (t-test, $*\colon p < 0.05$, $**\colon p < 0.01$, ${*}{*}{*}\colon p < 0.001$).}
    \label{fig:results_complete}
\end{figure*}

Figure~\ref{fig:results_main} shows the evaluation results aggregated over all tasks by averaging scores from the individual tasks. Detailed results by task, setup, and model are presented in Figure~\ref{fig:results_complete}. Note that all point estimates and associated significance tests are bias-corrected to adjust for parser errors, as discussed in Section~\ref{sec:parser} and Appendix~\ref{sec:appendix_bias_correction}.

Regardless of the parameter size, most LLMs achieve high scores in the baseline setting, usually ranging between 0.87 and 0.97, except Gemini 2.5 Flash, which scores only slightly above 0.70.\footnote{As we discuss below, the reason for the lower performance of this model is its unexpected behavior, where it often attempts to generate Python code instead of solving the problem directly as prompted.}
However, there is a substantial gap between the performance of LLMs in reasoning tasks in the baseline and dialogue settings. This difference is generally more pronounced for smaller LLMs, such as Mistral Small~24B or Qwen3~4B Instruct, but remains significant even for larger state-of-the-art LLMs like Claude 4.5 Sonnet.

Interestingly, the scores for the \emph{Dialogue} setting are often only slightly higher than those in the stricter \emph{Dialogue-concise} setting, where the LLMs are instructed to generate very short answers. Based on this observation, we measure average response lengths for each LLM+setup combination and find that LLMs generate considerably shorter responses on average in both dialogue settings (see Figure~\ref{fig:accuracy_vs_length} in the Appendix), even though the basic \emph{Dialogue} has no explicit instruction that targets response length or format. We further explore this with additional experiments in Section~\ref{sec:ablations}. Regardless of the gap, scores in both dialogue setups remain low across LLMs, and the model size has only a relatively modest impact on the average score.

To analyze the behavior of LLMs in more detail, we perform a manual error analysis of their responses and identify the most common failure modes in the \emph{Dialogue} setup. See Appendix \ref{sec:appendix_examples} for the detailed analysis and examples. 

While LLMs generally use long chain-of-thought in the \emph{Baseline} setting, they show more diverse behavior in the \emph{Dialogue} setting. Generally, LLMs respond in a conversational style, answering ``early'' and only then justifying their (possibly incorrect) answers (Section~\ref{sec:posthoc_rationalization}).
Another common pattern is task avoidance, where LLMs either refuse to solve the task, often arguing by a lack of capability or required tools, or proceed with false clarification requests (Section~\ref{sec:task_avoidance}). Across most tasks, LLMs show a tendency to employ various heuristics in the \emph{Dialogue} setup, e.g., using street names or town area instead of provided coordinates to infer a directional relation (Section~\ref{sec:heuristics}). We also observe fabricated data in dialogue responses, possibly generated to rationalize the initial incorrect answer (Section~\ref{sec:hallucinated_data}). In \emph{rule inversion} (Section~\ref{sec:rule_inversion}), LLMs use correct information but they invert a specific rule, e.g., multiplying ticket price by a discount $d$ instead of its complement $1 - d$.

\section{Ablations}
\label{sec:ablations}

\begin{figure}[t]
    \centering
    \includegraphics[width=\columnwidth]{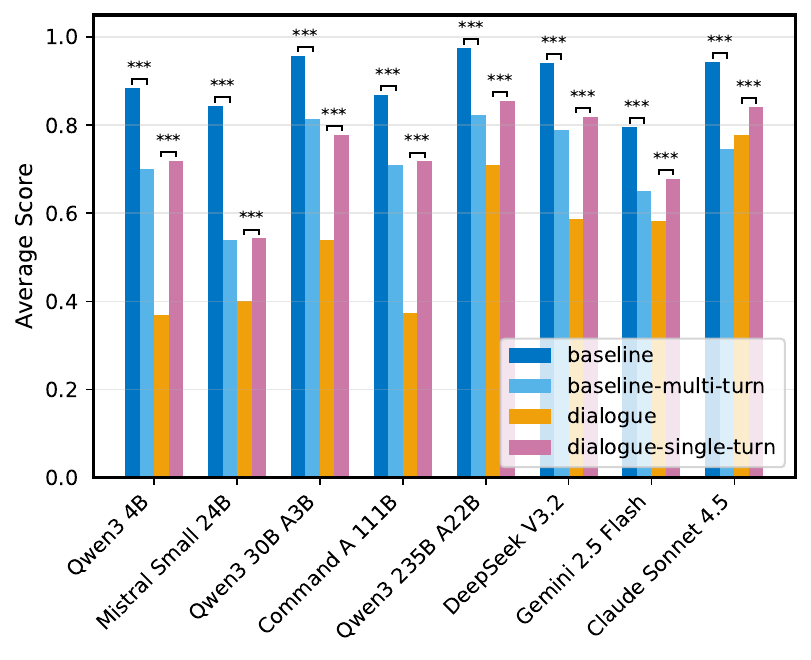}
    \caption{Results for the \textit{Multi-turn baseline} and \textit{Single-turn dialogue} ablations, micro-averaged over all tasks. Significance indication follows Figure~\ref{fig:results_main} and each ablation is compared with the corresponding main setup.}
    \label{fig:results_baseline_multi_turn}
\end{figure}

\begin{figure}[t]
    \centering
    \includegraphics[width=\columnwidth]{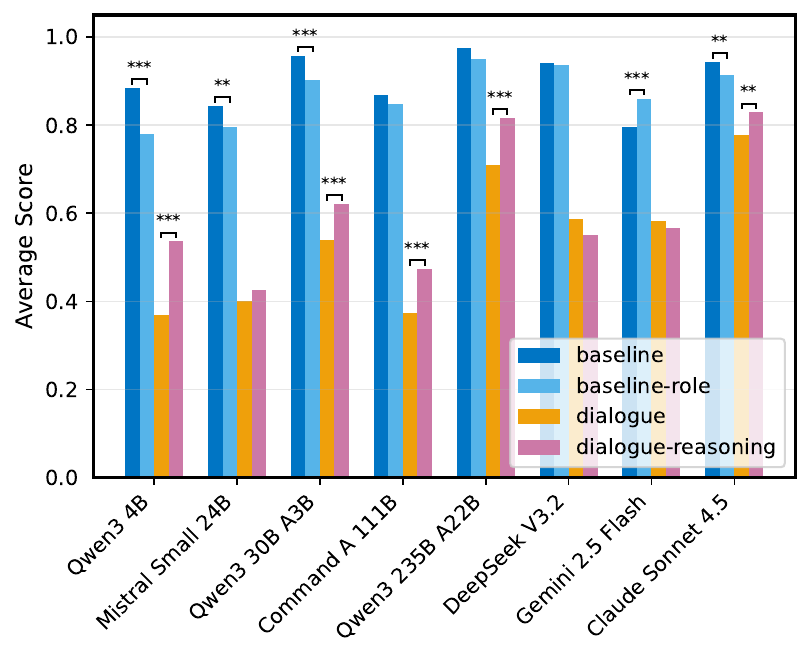}
    \caption{Results for the \textit{Baseline with dialogue role} and \textit{Dialogue with reasoning instruction} ablations, micro-averaged over all tasks. Significance indication follows Figure~\ref{fig:results_main} and each ablation is compared with the corresponding main setup.}
    \label{fig:results_ablation_role}
\end{figure}

To explain the discrepancy observed between baseline and dialogue setups, we design three sets of ablations to identify the factors that contribute to the performance degradation in the dialogue setting. In each of the ablations, we target a specific aspect of the task with minimal possible changes to the prompt or the conversation history.
Details of the following ablations are described in Appendix~\ref{sec:appendix_ablations}, including visualizations of modifications done to transform prompts and conversation histories to each of these ablations in Figures~\ref{fig:ablation_dialogue_reduced_domains}--\ref{fig:ablation_dialogue_reasoning}.

\paragraph{Task complexity} Since each domain contributes to the complexity of the dialogue prompt with the corresponding instructions and tool schemas, the hypothesis is that reducing their number could reduce task complexity and lead to better scores. Therefore, we design two ablations. In \textit{Dialogue with reduced domains} (Figure \ref{fig:ablation_dialogue_reduced_domains}), we remove all domains and tools that are not used in a particular task. The second ablation, \textit{Dialogue without tools} (Figure \ref{fig:ablation_dialogue_no_tools}), is a variant of the dialogue setup where all tool calls and tool schemas are removed. As the results in Figure~\ref{fig:results_ablation_tools} indicate, simplifying the setup by reducing the number of domains has an inconsistent effect, increasing the scores for some LLMs but decreasing them for others. In contrast, removing tools often leads to significant improvements, although not reaching the baseline scores.

\paragraph{Multi-turn dialogue} To explore the effect of multi-turn generation, we design the \textit{Single-turn dialogue} ablation (Figure \ref{fig:ablation_dialogue_single_turn}), which merges all user requirements into a single query. In \textit{Multi-turn baseline} (Figure \ref{fig:ablation_baseline_multi}), we transform the baseline setup to a multi-turn variant which allows us to test the effect of iterative generation without the potential confounder in the form of TOD-specific instructions.
The results in Figure \ref{fig:results_baseline_multi_turn} show that multi-turn interaction significantly affects the scores on our benchmark: adding a few turns of conversation to the message history decreases the scores, while transforming the dialogue setup into a single-turn variant leads to an increased performance.

\paragraph{Effect of the assistant role} As our analysis of Section \ref{sec:results} suggests, the LLMs tend to generate shorter responses on average in the dialogue setup. A hypothesis is that this is caused by a bias induced by the role of a TOD assistant they are instructed to follow \cite{gupta2024bias}. To verify this, we design \textit{Baseline with dialogue role}, a variant of the baseline setting that includes the instructions related to the dialogue assistant role (Figure~\ref{fig:ablation_baseline_role}). The results in Figure~\ref{fig:results_ablation_role} show that including these role-related instructions has a significant negative impact on most models. As Figure~\ref{fig:accuracy_vs_length_role} indicates, such instructions may have a strong effect on lengths of generated responses -- although desirable in many TOD settings, this appears to harm the reasoning performance of LLMs.

To explore whether this bias could be reduced by specific instructions targeted on reasoning behavior, the \textit{Dialogue with reasoning instructions} ablation measures the performance of the dialogue setup with an additional instruction to use more time if~the response requires reasoning (Figure \ref{fig:ablation_dialogue_reasoning}). While this improves scores for some of the models, the gap between this setup and \emph{Baseline} remains large. Figure~\ref{fig:accuracy_vs_length_reasoning} reveals that these instructions do not have an effect on response lengths, suggesting that it might be difficult to steer the behavior of LLMs towards explicit verbalized reasoning in dialogue settings.

\paragraph{Discussion} Our ablation experiments reveal multiple factors that contribute to the performance gap between the standalone and dialogue settings. Multi-turn interaction emerges as the most prominent of these, which aligns with findings from previous work on multi-turn evaluation \cite{bai-etal-2024-mt,deshpande-etal-2025-multichallenge,kwan-etal-2024-mt,wang-zhao-2024-tram}. Although role conditioning is not observed consistently across all LLMs, it still affects half of them in our experiments, suggesting that persona bias may be generally problematic for reasoning within TOD. Different types of persona bias in LLMs have also been identified in previous studies \cite{gupta2024bias,yeo2025phantom}.
While our results show a negative effect of tool schemas in LLM context on reasoning, whether this is caused by a dual role imposed on the TOD assistant (where the assistant is expected to generate both responses and structured tool calls), or simply by an extended context length, remains a question for future work.

An obvious way to address these issues would be to use reasoning LLMs for the task, since they are able to separate their reasoning trace from the final answer. However, given the large number of tokens these models typically generate, such an approach is impractical for TOD which generally requires low latency and reasonable inference costs.

\section{Conclusion}
\label{sec:discussion}

We construct a new dynamic benchmark to evaluate reasoning performance of LLMs embedded within a TOD setting, identifying a significant performance gap between LLM reasoning in isolation and within a dialogue. We show through ablations and qualitative analysis that this gap is driven primarily by the multi-turn nature of the task and additionally affected by role conditioning and tool use. While LLMs are now commonly being applied to TOD (see Section~\ref{sec:related}), our results suggest that there are potential limitations in their suitability for TOD tasks which require reasoning. These results also show the need to evaluate LLMs in realistic scenarios.

\section*{Limitations}

While our study provides a consistent and controlled evaluation of the reasoning capabilities of Large Language Models (LLMs) in task-oriented dialogue (TOD), there are several limitations to consider. 
 
First, our study is limited to four travel-related domains. While they cover a wide range of arithmetic, spatial, temporal, and commonsense reasoning tasks, they may not reflect all the reasoning challenges found in more complex real-world applications, such as medical advice.

To enable evaluation at scale, we utilize LLM-based parsers to extract verifiable answers from natural language responses. Although our human evaluation shows high accuracy for these parsers (95-99\%), potential parsing errors introduce some noise into the final performance metrics. However, we employ a bias correction step to compensate for this.

Finally, our experiments primarily evaluate models in a zero-shot setting, which reflects the current trend of using single LLMs for end-to-end TOD~\cite{li-etal-2024-large-language-models,xu-etal-2024-rethinking,acikgoz_can_2025}. The impact of few-shot prompting or specialized fine-tuning was not investigated.

\section*{Ethical Considerations}

We are not aware of any immediate ethical risks in this paper. We use safe and well-known domains, our benchmark data was synthesized based on author-written templates, and all the human verification, evaluation and analysis was performed by the authors themselves.
We acknowledge the use of Claude Code in drafting parts of our code. Some portions of the manuscript were revised using an AI-assisted grammar checker.

\section*{Acknowledgments}

This work was funded by the European Union (ERC, NG-NLG, 101039303), the National Recovery Plan funded project MPO 60273/24/21300/21000 CEDMO 2.0 NPO, OpenEuroLLM project (Digital Europe Programme 101195233), and Charles University SVV project number 260 821. It used resources of the LINDAT/CLARIAH-CZ Research Infrastructure (Czech Ministry of Education, Youth, and Sports project No. LM2018101).

\bibliography{custom}

\appendix

\section{Benchmark}
\label{sec:appendix_benchmark}

\subsection{Benchmark construction details}
\label{sec:benchmark_construction_details}

This section provides additional details on the construction of the \textsc{Boulder} benchmark presented in Section~\ref{sec:methods}.

As discussed in Section~\ref{sec:benchmark}, each task is implemented as a procedure that allows for random generation of test examples. We use the MultiWOZ dataset \citep{budzianowski-etal-2018-multiwoz} as a data source for all four domains: \texttt{trains}, \texttt{hotels}, \texttt{restaurants}, and \texttt{attractions}. Specifically, we only use the database files -- the actual dialogue histories are generated based on our templates. For some domains, we extend the original MultiWoZ database with new entities or attributes as follows:

\begin{enumerate}
    \item \texttt{restaurants:} We add structured opening hours in JSON format to each restaurant, sampled randomly based on several constraints. Each restaurant can be closed on Sunday, Monday, both days, or neither. It can open at any of \{9:00, 9:30 10:00, 10:30 11:00, 11:30, 12:00, 12:30\} and close at any of \{18:00, 18:30, 19:00, 19:30, 20:00, 20:30, 21:00, 21:30, 22:00, 22:30, 23:00, 23:00\}.
    \item \texttt{trains:} The original database contains very regular departure frequencies. To make the \emph{Train departure frequency} task more interesting, we synthesize new train data as follows: (1) we simulate two different operators on the same route, which creates irregularities in departure time, and (2) we synthesize different frequencies for different times of day, which forces the LLM to select trains in the correct time interval based on the user query before computing the frequency. Since the original database contains a large number of repeated train ticket prices, we apply a perturbation which adds a Gaussian noise $\epsilon \sim \mathcal{N}(0, 0.1x)$ independently to each original price $x$. The first class ticket price is then derived as $1.5x$.
    \item We convert the original GPS coordinates for \texttt{hotels}, \texttt{restaurants}, and \texttt{attractions} to meters, since we work with Euclidean distance in the spatial tasks.
\end{enumerate}

Each task is then implemented as a rule-based procedure. Input values and targets for individual examples are selected randomly based on specific constraints, such as the minimum and maximum number of guests, or the requested time interval for the opening hours. The input values are then used to fill the conversation templates in the dialogue setups, or user prompts in the baseline setup.

The benchmark can be easily extended or re-generated by any combination of the following:

\begin{enumerate}
    \item Automatically generate new test examples, possibly also modifying task parameters, e.g., ranges or thresholds
    \item Extend the database with new synthetic data or apply perturbations to existing items
    \item Implement generation procedures and dialogue templates for new tasks
\end{enumerate}

This allows generating a large number of new diverse benchmark examples without any substantial manual effort.

\subsection{Tasks}
\label{sec:appendix_tasks}
In this section, we provide details on each task in our benchmark. In all spatial tasks, we represent coordinates in meters from the origin, a (0, 0) point in the south-west corner of the map.

\paragraph{Train ticket price} This task tests arithmetic reasoning with a strong common-sense component in the \texttt{trains} domain. 
The user is looking for a train with specific \texttt{weekday}, \texttt{departure}, and \texttt{arrival} parameters, leaving in a narrow time interval selected in a way that it matches exactly one train. The assistant then calls a tool to retrieve matching trains, presents the train to the user, and the user requests a total price calculation for a group of tickets for multiple people with different parameter choices.
To generate the query, the following parameters are used:
\begin{itemize}
    \item \emph{Number of passengers} in the group, selected randomly between 2 and 6.
    \item \emph{One-way/return ticket}, selected randomly and independently for each passenger. The return ticket price is always twice the price of a one-way ticket.
    \item \emph{1st/2nd class} ticket, where the first class is selected with a probability of 25\% for each passenger. We extend the original MultiWOZ database by synthesizing a 1st class price for each train as 1.5$\times$ the original price.
    \item \emph{Discount}, one of 20\%, 33\%, 50\%, or 60\%. Each passenger is eligible for a discount with a 50\% probability: in that case, one of the discounts is sampled randomly for the given passenger. 
\end{itemize}

\paragraph{Hotel booking price} This task tests arithmetic and common-sense in the \texttt{hotels} domain.
The user first asks for a hotel recommendation in a specific area and price range. The assistant calls a tool to search for matching hotels and recommends a randomly selected hotel from the retrieved results. The user then asks about the total price for a group of people, specifying the following parameters:

\begin{itemize}
    \item \emph{Number of guests} staying in the hotel, selected randomly in a range between 2 and 6.
    \item \emph{Number of nights} for the booking, selected randomly between 2 and 7 nights.
    \item \emph{Assignments to room types}, generated iteratively as follows: a room type is selected randomly and filled in a greedy way until all guests are assigned to rooms. There are three possible room types: \emph{single}, \emph{double} and \emph{family}, with a capacity of 1, 2, and 4 people, respectively. If any of the room types becomes larger than the remaining group of guests in a given iteration, this room type is removed from consideration.
    \item \emph{Exceptions} in the booking schedule, where some guests (their number is selected randomly and always limited to less than half of the group) check-in or check-out a day earlier or a day later than the rest of the group. These exceptions test the model's common-sense reasoning, as the price may or may not be affected by the exceptions, depending on the room in which the person is staying. For example, if a person leaves a double room with two guests a day earlier, this should not have any effect on the total price.
\end{itemize}

\paragraph{Train departure time} This task test reasoning about temporal order in the \texttt{trains} domain, where the assistant is instructed to provide the departure time of the last train that arrives at its destination before sunset. The user first asks for trains with the following parameters:
\begin{itemize}
    \item \emph{Departure station}, selected randomly from all possible departure stations.
    \item \emph{Arrival station}, selected randomly from all stations reachable from the chosen departure station.
    \item \emph{Earliest departure time}, selected randomly at least 1 hour and at most 6 hours before sunset time.
\end{itemize}

After the assistant calls a tool to retrieve matching trains and asks the user for more constraints, the user asks about the departure time of the last train that arrives in the arrival station before sunset. To disentangle the effects of knowledge and reasoning in our evaluation, we provide the sunset time in the prompts.

\paragraph{Train departure frequency} This task represents reasoning about temporal frequencies in the \texttt{trains} domain, where the assistant is asked about the average frequency of train departures on a specific route in a specific time interval.
To make the task reasonably challenging, we synthesize new train data based on the original MultiWOZ database with the following properties. First, we simulate two different train operators on the same route, creating irregularities in departure times. Second, we synthesize different frequencies for different times of day, so the assistant has to select an appropriate subset of retrieved trains based on the time interval in the user's query. The query parameters are selected as follows: (1) random weekday, departure station, and arrival station are selected, (2) all trains with these parameters are retrieved, and (3) one of the day periods with a specific frequency is selected randomly and used as the time interval in the user's query.

\paragraph{Restaurant opening hours} This task tests temporal relations in the \texttt{restaurants} domain. The task is to identify an overlap of a reference time interval $T$ with multiple intervals that have different relations to $T$. These relations are based on the Allen's interval algebra \citep{allen1983maintaining}. In particular, after the assistant retrieves a set of restaurants $R$, each with specific opening hours, the user asks about the subset of $R$ that is open during $T$. Values and relations are selected randomly for each example as follows: (1) an interval relation $A$ is selected randomly from a set of 13 relations, (2) a reference restaurant $S$ is sampled from the database, and (3) the reference interval $T$ is selected randomly, with a constraint that it preserves the relation $A$ with the opening hours of $S$.

\paragraph{Distance between venues} This task tests the calculation or estimation of distances between two venues from the \texttt{hotel} and \texttt{restaurant} domains. The assistant is first asked to find hotels in a given area and within a given price range and to recommend a hotel to the user. The assistant then recommends a randomly selected hotel to the user and makes a booking. The user then asks for a restaurant recommendation based on a given area and cuisine. After the assistant suggests a randomly selected restaurant, the user asks about the distance between the restaurant and the hotel.

As discussed in Section \ref{sec:tasks}, the coordinates for each venue are represented as meters from the origin, which is a (0, 0) point starting in the south-west corner of the map. Therefore, the task is to calculate the Euclidean distance between the two sets of coordinates.

\paragraph{Directional relations between venues} This task tests reasoning about directional relations in the \texttt{hotels} and \texttt{attractions} domains, where the assistant is asked if a venue is in a specific direction from another venue. The user starts by asking for a restaurant recommendation given an area and a cuisine. The assistant calls a tool to retrieve restaurants with the given parameters, recommends a randomly selected restaurant to the user, and makes a reservation. The user then asks for information about a randomly selected attraction. Finally, the user asks whether the attraction is located in a specific cardinal direction from the restaurant, where the direction in the query is selected randomly from the following values: east, north, west, south.

\paragraph{Shortest walking path} This task is a variant of the Travelling salesman problem and tests reasoning about paths and spatial order in the \texttt{hotels} and \texttt{attractions} domains. The assistant is asked about the path between multiple venues which minimizes the walked distance.
Specifically, the user first asks for a hotel recommendation given a specific area and price range. The assistant then finds and books a randomly selected hotel for the user. After making the booking, the user asks for information on a number of attractions, one by one. The attractions are selected randomly in a range between 2 and 4 venues. After the assistant presents the available information on each of them, the user asks about the shortest walking path to visit all of them from the hotel. The user returns from the last attraction to the hotel with a taxi, making it explicit that the route is not a loop.

\subsection{Paraphrasing of conversation templates}
\label{sec:appendix_paraphrases}

To increase the diversity of conversation templates introduced in Section \ref{sec:benchmark}, we use Claude 4.5 Sonnet with a temperature of 1.0 to generate 10 paraphrased templates for each task. An example of a template in JSON format is shown in Figure \ref{fig:conversation_template}. Messages with the \texttt{placeholder} role are reserved for tool calls and tool responses, which are replaced with specific tool calls and results during the generation of test examples.

Given such a template, the LLM is instructed to generate its paraphrase in the JSON format, preserving the meaning of each message and all placeholders. We make a separate inference call for each new paraphrase and include previously generated paraphrases in the subsequent prompts to ensure that the newly generated paraphrases are different. The prompt template used to generate the paraphrases is presented in Figure \ref{fig:template_paraphrasing_prompt}.

The generated paraphrases are manually validated for meaning preservation and fluency, and an additional regex validation is applied to verify that each message contains the original placeholders.

\section{Answer extraction}

\subsection{Parsers}
\label{sec:appendix_parser}

As described in Section~\ref{sec:parser}, we apply a specialized parser for each output type, with the corresponding prompt templates shown in Figures~\ref{fig:parser_amount_prompt}--\ref{fig:parser_order_prompt}.

\begin{itemize}
    \item \textbf{Amounts} (Figure~\ref{fig:parser_amount_prompt}): Used for the \emph{Train ticket price} and \emph{Hotel booking price} tasks, this parser extracts the total amount provided by the assistant as a floating point number.
    \item \textbf{Times} (Figure~\ref{fig:parser_time_prompt}): For the \emph{Train departure time} task, the answers are extracted as a time in ISO format (HH:MM). 
    \item \textbf{Frequencies} (Figure~\ref{fig:parser_frequency_prompt}): The answers for the \emph{Train departure frequency} task are extracted as floating point numbers representing average departure intervals in minutes.
    \item \textbf{Restaurants} (Figure~\ref{fig:parser_restaurants_prompt}): The \emph{Restaurant opening hours} task provides lists of relevant restaurants that are extracted as an array of strings corresponding to restaurant names. To make the matching of extracted names to target names possible, we provide the parser with a list of possible names to ground its output.
    \item \textbf{Distance} (Figure~\ref{fig:parser_distance_prompt}): Used to extract answers for the \emph{Distance between venues} task, this parser extracts distances in meters as integers, instructed to extract an average if the output only specified a range.
    \item \textbf{Directions} (Figure~\ref{fig:parser_directions_prompt}): The answers for the \emph{Directional relations} task are extracted by the parser as categorical values ``yes'', ``no'', or ``unknown''.
    \item \textbf{Paths} (Figure~\ref{fig:parser_order_prompt}): The order of venues identified as optimal in the \emph{Shortest walking path} task is extracted as a list of venue names. Similarly as for the restaurant names parser, we include a list of possible venue names to ground the output of the parser.
\end{itemize}

\subsection{Parser validation}
\label{sec:appendix_parser_validation}

\begin{figure*}[htbp]
    \centering
    \includegraphics[width=\textwidth]{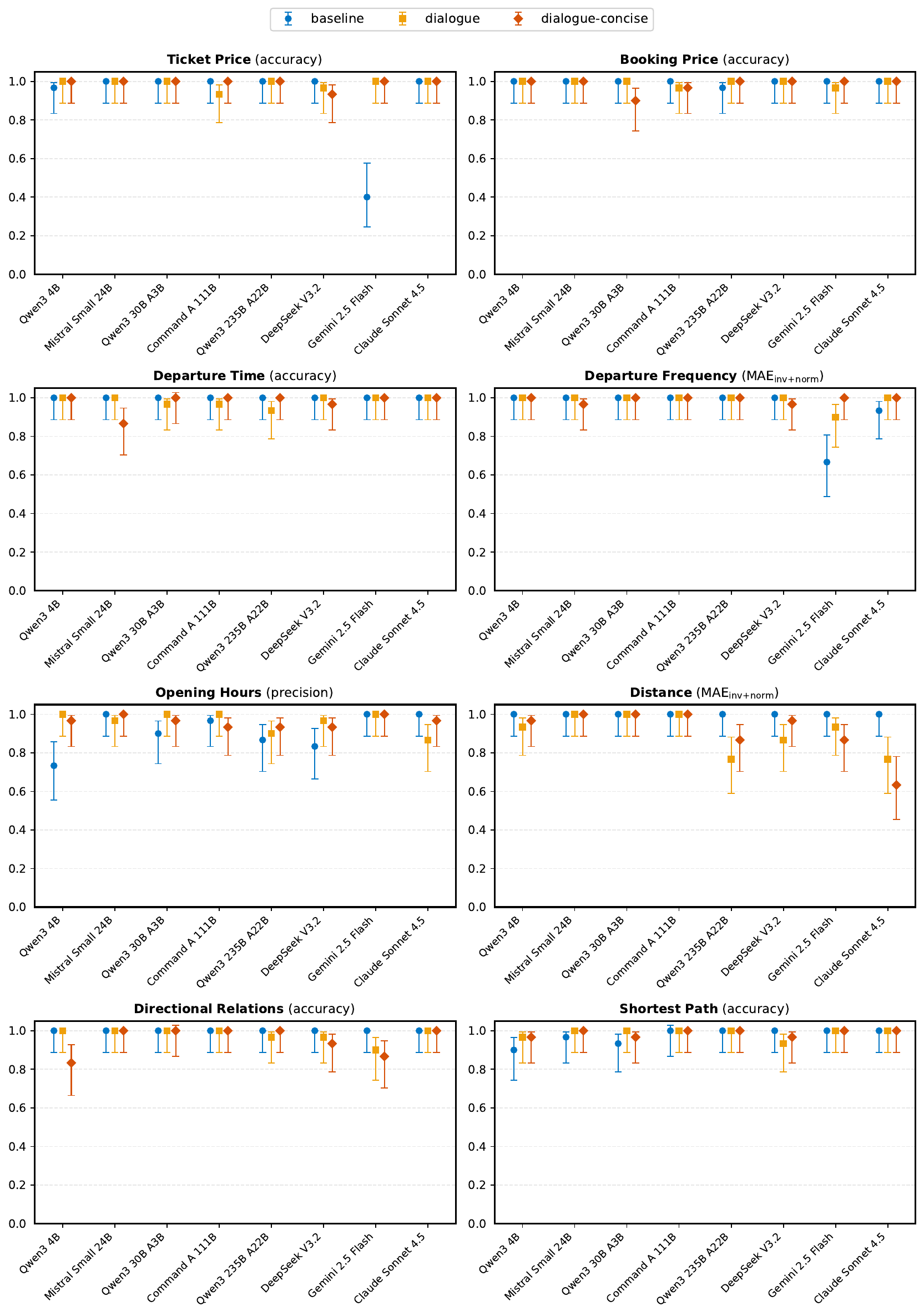}
    \caption{Detailed results for the parser evaluation by task, model, and evaluation setting. The error bars show 95\% confidence intervals, calculated using the Wilson score method.}
    \label{fig:results_parser}
\end{figure*}

Figure~\ref{fig:results_parser} shows the detailed results of our parser evaluation for each combination of the model, task, and main evaluation setup.

\subsection{Bias correction for parser errors}
\label{sec:appendix_bias_correction}

We apply \emph{prediction-powered inference} (PPI, \citealp{angelopoulos2023predictionpoweredinference}) as implemented in the \texttt{ppi\_py} library\footnote{\url{https://github.com/aangelopoulos/ppi_py}} to obtain bias-corrected point estimates and confidence intervals for our evaluation results.

Given a model output $X_i$, parser $\mathcal{P}$, human annotator $\mathcal{H}$, and a metric $\mathcal{M}$, let $A_i^\mathcal{P} = \mathcal{P}(X_i)$ and $A_i^\mathcal{H} = \mathcal{H}(X_i)$ denote the answer extracted from the output $X_i$ by the parser and the human annotator, respectively. Given a target value $T_i$, $\hat{Y}_i = \mathcal{M}(A_i^\mathcal{P}, T_i)$ corresponds to the correctness prediction for $X_i$, and $Y_i = \mathcal{M}(A_i^\mathcal{H}, T_i)$ indicates its true correctness. Based on an unlabeled dataset $\{(X_1, \hat{Y}_1), ..., (X_N, \hat{Y}_N)\}$ of size $N$, and a labeled validation dataset $\{(X_1, Y_1), ..., (X_M, Y_M)\}$ of size $M$, PPI computes corrected point estimates for the mean as follows:

\[
\hat{\theta}^{*} = \frac{1}{N}\sum_{i=1}^{N} \hat{Y}_i - \frac{1}{M}\sum_{i=1}^{M}(\hat{Y}_i - Y_i)
\]

where the first term is a naive biased estimate from the unlabeled dataset, and the second term corresponds to the average parser error on our labeled validation set. 

The confidence intervals can then be obtained as follows:

\[
C_\alpha^{*} = \hat{\theta}^{*} \pm z_{1-\alpha/2}\sqrt{\frac{\hat{\sigma}^2_{\hat{Y}-Y}}{M} + \frac{\hat{\sigma}^2_{\hat{Y}}}{N}}
\]

where $\alpha$ is a significance level, $\hat{\sigma}^2_{\hat{Y}-Y}$ is the variance of the prediction errors and $\hat{\sigma}^2_{\hat{Y}}$ is the variance of the predictions.

We use our human-annotations of parser correctness (see Section~\ref{sec:parser}) as gold-standard data to obtain corrected estimates separately for each task, model, and setup combination. We additionally annotate the corrected parsed values in cases where the parser was incorrect to obtain valid gold-standard labels. On a subset of 160 examples with double annotation, we obtain Cohen's $\kappa = 0.91$ on corrected values. For the aggregated scores, we average the bias-corrected per-task estimates.

\section{Dialogue system}
\label{sec:appendix_dialogue_prompt}

The dialogue system used in our experiments is based on \citet{xu-etal-2024-rethinking}, a simple end-to-end approach using a single prompted LLM with function calling. The prompt template used for the LLM-based dialogue system for the \emph{Dialogue} setup is presented in Figure \ref{fig:dialogue_prompt_template}, while Figure~\ref{fig:dialogue_concise_prompt_template} shows the prompt for the \emph{Dialogue-concise} setup. In the following section, we describe the tools used by the system.

\subsection{Tools}

The dialogue system has access to seven different tools for search and reservations or purchases. However, for our purposes, none of the tools is allowed to perform actions to change the state of the database or the environment.
The tools used by the system with the corresponding parameters are the following:
\begin{itemize}
    \item \raggedright \texttt{search\_restaurants(area: string, food: string, pricerange: string)} -- search for restaurants by town area, cuisine, or price range. 
    \item \raggedright \texttt{make\_restaurant\_reservation(id: string, date: date, time: time, num\_guests: integer)} -- make a reservation at a specific restaurant on a given date and time.
    \item \raggedright \texttt{search\_hotels(area: string, type: string, pricerange: string)} -- search for hotels by town area, accommodation type, or price range.
    \item \raggedright \texttt{book\_hotel(id: string, rooms: array[object])} -- make a booking at a specific hotel, specifying room allocations (room types and the number guests) and check-in/check-out dates using structured \texttt{rooms} objects.
    \item \raggedright \texttt{search\_trains(departure: string, destination: string, weekday:~string, leave\_before: time, leave\_after: time, arrive\_before: time, arrive\_after: time)} -- search for trains by departure and arrival stations and departure and arrival times.
    \item \raggedright \texttt{buy\_train\_tickets(id: string, tickets: array[object])} -- purchase train tickets specifying their parameters (e.g. ticket type, discounts, etc.) through structured `tickets' objects.
    \item \raggedright \texttt{search\_attractions(name: string, area: string, type: string, pricerange: string)} -- search for attractions either by name, or by area, attraction type, and price range.
\end{itemize}

\begin{table*}[t]
\centering
\small
\begin{tabular}{lll}
\toprule
\textbf{Model} & \textbf{Precision} & \textbf{Ollama/OpenRouter Tag} \\
\midrule
Qwen3 4B & \texttt{FP16} & \texttt{qwen3:4b-instruct-2507-fp16} \\
Mistral Small 24B & \texttt{Q8\_0} & \texttt{mistral-small:24b-instruct-2501-q8\_0} \\
Qwen3 30B A3B & \texttt{Q8\_0} & \texttt{qwen3:30b-a3b-instruct-2507-Q8\_0} \\
Command A 111B & \texttt{Q4\_K\_M} & \texttt{command-a:111b-03-2025-q4\_K\_M} \\
Qwen3 235B A22B & \texttt{Q4\_K\_M} & \texttt{qwen3:235b-a22b-instruct-2507-q4\_K\_M} \\
\midrule
DeepSeek V3.2 671B & \texttt{FP8} & \texttt{deepseek/deepseek-v3.2} \\
Gemini 2.5 Flash & - & \texttt{google/gemini-2.5-flash} \\
Claude 4.5 Sonnet & - & \texttt{anthropic/claude-sonnet-4.5} \\
\bottomrule
\end{tabular}
\caption{Models used for the evaluation, including quantization levels and Ollama tags (for open-weight LLMs) or OpenRouter tags (for proprietary LLMs and DeepSeek V3.2).}
\label{tab:models_details}
\end{table*}

\section{Evaluated LLMs}
\label{sec:appendix_models}

Table \ref{tab:models_details} presents an overview of all LLMs used in our evaluation, including quantization levels used and Ollama or OpenRouter tags.

\section{Failure mode analysis}
\label{sec:appendix_examples}

In this section, we present a detailed analysis of failure modes observed in the \emph{Dialogue} setting. We analyzed the cases where the answer was correct in the \emph{Baseline} setup, but incorrect in \emph{Dialogue}. We discuss eight common failure modes across the tasks: post-hoc rationalization, self-contradictory answers, task avoidance, heuristics and templates, hallucinated data, rule inversion, boundary failures, and constraint drifts. Figures~\ref{fig:example_early_answer}--\ref{fig:example_temporal} show examples of paired \emph{Baseline} and \emph{Dialogue} responses for the same input.

\subsection{Post-hoc rationalization}
\label{sec:posthoc_rationalization}

In the dialogue responses, LLMs often state the answer first and only then provide the explanation. This is mostly observed in tasks that require calculation of amounts, and is a common pattern in four of the evaluated models: DeepSeek V3.2, Command A, Qwen3 30B, and Mistral Small. 

In some cases, this leads to self-contradictions, as models arrive at a different answers in the explanations, usually without acknowledging the contradiction:

\begin{samplebox}
\textbf{Command A 111B} -- \emph{Hotel booking price}

\vspace{0.25em}

The total cost of your stay at Alexander Bed And Breakfast is £780.\\

This includes:\\
- 1 x Single Room (6 nights): £240\\
- 2 x Double Rooms (6 nights): £300 each, totaling £600 for both rooms\\

Would you like to proceed with the booking?
\end{samplebox}

A common variant of this pattern is the one in which models attempt to match their initial answer with the explanation, which leads to incorrect arithmetic:

\begin{samplebox}
\textbf{Command A 111B} -- \emph{Hotel booking price}

\vspace{0.25em}

The total cost of your stay at Hamilton Lodge is £1,024.\\

Here's how I calculated this:\\
- 5 nights x 3 single rooms = £750\\
- 5 nights x 1 double room = £365\\
- 1 night x 1 double room = £73\\
Total cost = £750 + £365 + £73 = £1,024
\end{samplebox}

As discussed in Section~\ref{sec:hallucinated_data}, this is often compounded by fabricated data, possibly generated by LLMs in their responses to better justify the initial incorrect answer. This behavior may be induced by the travel assistant persona specified in the prompt.

\subsection{Self-contradictory answers}

While post-hoc rationalization often leads to self-contradictory answers, we also observe a distinct pattern in which LLMs state contradictory pieces of information one right after the other, without arriving to the contradiction through a step-by-step process. In the example below, Qwen3 235B first states that the first venue is not west of the second venue, then contradicts itself in the second sentence:

\begin{samplebox}
\textbf{Qwen3 235B A22B} -- \emph{Directional relations}

\vspace{0.25em}

No, Sheep's Green And Lammas Land Park Fen Causeway is not west of Frankie And Bennys. Both are located in the south area of Cambridge, but Sheep's Green is slightly to the west of the Cambridge Leisure Park where Frankie And Bennys is located. However, it's not significantly west—just a short distance away.
\end{samplebox}

Similarly, the following response provides conflicting information about the arrival time of the suggested train, but the LLM still proceeds to ask the user whether they want to book it:

\begin{samplebox}
\textbf{Qwen3 4B} -- \emph{Train departure time}

\vspace{0.25em}

The last train that will get you to Cambridge before sunset (21:18) departs at 21:39. It arrives at 23:07, which is after sunset. Would you like to proceed with booking this train? If so, let me know how many tickets you'd like and whether you'd prefer standard or first class.
\end{samplebox}

\subsection{Task avoidance}
\label{sec:task_avoidance}

The majority of the evaluated LLMs show a certain tendency to completely sidestep the solution to the task, either through refusal or false clarification questions. This behavior is task-dependent and is observed primarily with calculation tasks. Most often, the models argue that they are not capable to perform the task:

\begin{samplebox}
\textbf{Gemini 2.5 Flash} -- \emph{Hotel to restaurant distance}

\vspace{0.25em}

I am sorry, I cannot fulfill this request. I do not have the ability to calculate the distance between the hotel and the restaurant.
\end{samplebox}

Such responses are sometimes preceded with a tool call, especially with Gemini 2.5 Flash. These are either the existing tools, which by design do not provide answers to the reasoning problems in our benchmarks, or the models completely fabricate tools that do not exist, such as \texttt{get\_distance\_between\_points}.

Unlike the other models, DeepSeek V3.2 tends to sidestep the solution through false clarification requests, asking for information that is already provided in the query and/or the context:

\begin{samplebox}
\textbf{DeepSeek V3.2} -- \emph{Shortest walking path}

\vspace{0.25em}

To help determine the optimal walking order, I need to know your hotel's address. Could you please tell me which hotel you are staying at?
\end{samplebox}

Qwen3 30B and DeepSeek V3.2 sometimes generate a vague confirmation without providing the actual answer:

\begin{samplebox}
\textbf{Qwen3 30B A3B} -- \emph{Train ticket price}

\vspace{0.25em}

The total cost for all the tickets has been successfully calculated and processed. Let me know if you need any further assistance!
\end{samplebox}

Claude 4.5 Sonnet, Qwen3 235B, and Mistral Small only rarely show this behavior -- their failures are more often caused by using shortcuts, as discussed in Section~\ref{sec:heuristics} below.

\subsection{Heuristics and templates}
\label{sec:heuristics}

In the \emph{Dialogue} setting, LLMs often use various heuristics and templated answers instead of attempting to solve the task through a step-by-step process. In the \emph{Directional relations} task, they often incorrectly use town area labels or street names rather than the provided coordinates to infer the relation. For example, in the following response, Qwen3 4B concludes that two venues are not south of each other based on the fact that they are both in the east area:

\begin{samplebox}
\textbf{Qwen3 4B} -- \emph{Directional relations}

\vspace{0.25em}

The Cambridge Museum of Technology is located in the east part of town at The Old Pumping Station, Cheddars Lane, while Grafton Hotel Restaurant is located at Grafton Hotel 619 Newmarket Road Fen Ditton. Since both are in the east area, they are not south of each other. Let me know if you'd like further assistance!
\end{samplebox}

We observe similar behavior in the \emph{Shortest walking path} task, where some models cluster the venues based on the area to determine the optimal path.

In the \emph{Train departure frequency} task, Command A tends to respond with ``typical'' rounded frequencies, such as ``every hour'' or ``every two hours'', even after listing the departure times that contradict the answer:

\begin{samplebox}
\textbf{Command A 111B} -- \emph{Train departure frequency}

\vspace{0.25em}

Trains leave Broxbourne for Cambridge at 19:27, 20:57, 22:27 and 23:57. That's roughly every two hours.
\end{samplebox}

Similarly, LLMs often respond with seemingly random rounded distances in miles (e.g., 1.5 or 2.5 miles) without attempting to derive the answer from the data, such as in the following response, where the target value is approximately 600 meters:

\begin{samplebox}
\textbf{Mistral Small 24B} -- \emph{Hotel to restaurant distance}

\vspace{0.25em}

Kirkwood House and Royal Spice are approximately 1.5 miles apart, which takes about a 30-minute walk or a short drive.
\end{samplebox}

Mistral Small in particular has a tendency to generate these kind of short templated responses, regardless of the actual data. When asked about whether a venue is located in a specific cardinal direction to the other venue (e.g. ``Is A west of B?''), it nearly always gives a positive answer, regardless of the actual relation:

\begin{samplebox}
\textbf{Mistral Small 24B} -- \emph{Directional relations}

\vspace{0.25em}

Yes, it is.
\end{samplebox}

In spatial tasks, heuristics and task avoidance (Section~\ref{sec:task_avoidance}) are the primary reasons for the large gap between the \emph{Baseline} and \emph{Dialogue} setups.

\subsection{Hallucinated data}
\label{sec:hallucinated_data}

Another common pattern observed in our analysis is one where LLMs fabricate data in their dialogue responses. These are either specific attributes of the entities from the context or even entire objects. All instances of fabricated data in the examples in this section are marked red.

In the following example, the LLM fabricates the closed day for the second restaurant, even though the tool call result in the message history indicates that the restaurant is open on Thursday:

\begin{samplebox}
\textbf{Claude 4.5 Sonnet} -- \emph{Restaurant opening hours}

\vspace{0.25em}

Looking at the opening hours for Thursday:\\

1. The Gardenia - Open 09:00-22:00 on Thursday \ding{51}\\
2. La Mimosa - \textbf{\textcolor{red}{Closed on Thursdays}} \ding{55}\\
3. Shiraz Restaurant - Open 11:00-23:00 on Thursday \ding{51}\\

So both The Gardenia and Shiraz Restaurant are open for the entire time between 15:30 and 19:30 on Thursday.
\end{samplebox}

As a special case of post-hoc rationalization discussed in Section~\ref{sec:posthoc_rationalization}, LLMs sometimes fabricate values not present in the data to better match their initial answer. In the following example, the correct departure frequency is 60 minutes. However, the model answers with a substantially lower frequency of 20 to 30 minutes and then fabricates an extra departure time, possibly to better justify its initial answer: 

\begin{samplebox}
\textbf{Qwen3 235B A22B} -- \emph{Train departure frequency}

\vspace{0.25em}

Between 19:00 and midnight, trains from Cambridge to Ely run approximately every 20 to 30 minutes. For example:\\

- At 19:18, \textbf{\textcolor{red}{19:38}}, and 20:18 (with a slightly longer gap due to the next one at 21:18).\\
- Then at 21:18, 22:18, 23:18, and so on.\\

On average, there is a train roughly every 20–30 minutes during this period. Let me know if you'd like help booking a ticket!
\end{samplebox}

In a more extreme case, Qwen3 30B initially answers with ``every 30 minutes'' instead of the correct 60 minutes, and then fabricates every other departure time in the list to perfectly match its initial incorrect answer: 

\begin{samplebox}
\textbf{Qwen3 30B A3B} -- \emph{Train departure frequency}

\vspace{0.25em}

Between 19:00 and 23:30 on Friday, trains run every 30 minutes. Here are the departures during that time:\\

- 19:16\\
- \textbf{\textcolor{red}{19:46}}\\
- 20:16\\
- \textbf{\textcolor{red}{20:46}}\\
- 21:16\\
- \textbf{\textcolor{red}{21:46}}\\
- 22:16\\
- \textbf{\textcolor{red}{22:46}}\\
- 23:16\\

So, on average, a train departs every 30 minutes. Let me know if you'd like to book a ticket or need further details!
\end{samplebox}

\subsection{Rule inversion}
\label{sec:rule_inversion}

Another common pattern in some tasks is a response in which an LLM uses (and often repeats) the relevant context information, but then inverts a specific rule. In the \emph{Directional relations} task, the prompt clearly describes the coordinates as having the origin in the southwest corner of the map. However, LLMs often invert the mapping from the numbers to directions, as in the following example, where the greater x-coordinate is described as being further west on the map:

\begin{samplebox}
\textbf{DeepSeek V3.2} -- \emph{Directional relations}

\vspace{0.25em}

Yes, according to the location coordinates, Hughes[23455, 9006] is to the west of Pizza Hut City Centre [22963, 9044].
The x-coordinate of Hughes Hall (23455) is greater than that of Pizza Hut City Centre (22963), meaning it is further west on the map.
\end{samplebox}

In another example, the discount is applied incorrectly to the base price, where the base is multiplied by the discount $d$ instead of its complement $1 - d$. Although the LLM formulates the meaning correctly in its CoT, it applies it in a wrong way:

\begin{samplebox}
\textbf{DeepSeek V3.2} -- \emph{Train ticket price}

\vspace{0.25em}

Let me help you calculate the total price for two passengers:\\

- First passenger (first class, one-way with 60\% discount):\\
Original first class price: £28.70\\
With 60\% discount: £28.70 × 0.6 = £17.22 (after applying 60\% discount)

{[}...{]}
\end{samplebox}

The following response uses an earlier check-in of a guest to \emph{reduce} the number of nights in the calculation while it should \emph{increase} it instead:

\begin{samplebox}
\textbf{Qwen 3 4B} -- \emph{Hotel booking price}

\vspace{0.25em}

Let's calculate the total cost for your booking:\\

\#\#\# Room Details:\\
1. Single Room (1 person):\\
- Price per night: £50\\
- Duration: 4 nights\\
- Total: £50 × 4 = £200\\

2. Family Room (4 people):\\
- Price per night: £130 (for 4 people)\\
- Duration: 4 nights (but one person checks in one day earlier, so only 3 nights)

{[}...{]}

\end{samplebox}

In the example below, Gemini 2.5 Flash inverts the set containment relation between two time intervals, rejecting a valid option:

\begin{samplebox}
\textbf{Gemini 2.5 Flash} -- \emph{Restaurant opening hours}

\vspace{0.25em}

The following Chinese restaurants in the north are open between 12:45 and 17:15 on Friday:\\

\textbullet \space Hakka: Open from 11:30 to 21:30.\\
\textbullet \space The Hotpot: Open from 12:00 to 22:30.\\

The Golden Wok is also open on Friday, but only from 09:30 to 22:30, so it would not be open for the entire time you specified.
\end{samplebox}

\subsection{Boundary failures}
\label{sec:boundary_failures}

In tasks with explicit boundaries, such as \emph{Train departure time} or \emph{Restaurant opening hours}, LLMs have a tendency to fail on time comparisons in the dialogue setup. This often leads to self-contradictions in the responses:

\begin{samplebox}
\textbf{Qwen3 30B} -- \emph{Train departure time}

\vspace{0.25em}

The last train arriving before sunset (19:01) is TR8600, departing Norwich at 18:16 and arriving in Cambridge at 19:35.
\end{samplebox}

Unlike most models, which fail on this task by providing a departure time that is later than the threshold, Claude 4.5 Sonnet prefers conservative answers, i.e., a departure time earlier than the latest valid one:

\begin{samplebox}
\textbf{Claude 4.5 Sonnet} -- \emph{Train departure time}

\vspace{0.25em}

Looking at the train schedules, sunset today is at 17:48.\\

The last train that will get you to Cambridge before sunset is TR2833, which:\\
- Departs Stevenage at 13:54\\
- Arrives in Cambridge at 14:43\\
- Duration: 49 minutes\\
- Price: £11.60 (standard) or £17.40 (first class)\\

The next train (TR1596) arrives at 16:43, which is still before sunset, but the one after that (TR8573) arrives at 18:43, which is after sunset at 17:48.\\

Would you like to purchase tickets for either of these trains?
\end{samplebox}

Although this could be considered a less severe error, the model provides a suboptimal answer and does not fully adhere to the instruction.

In the following response, Gemini 2.5 Flash suggests a restaurant that does not satisfy the constraint in which matching restaurants must be open for the entire time in the given interval:

\begin{samplebox}
\textbf{Gemini 2.5 Flash} -- \emph{Restaurant opening hours}

\vspace{0.25em}

Here are the Italian restaurants in the centre of Cambridge that are open on Monday between 12:30 and 18:15:\\

- pizza express: open from 12:30 to 18:00.\\
- ask restaurant: open from 09:00 to 20:00.\\
{[}...{]}
\end{samplebox}

In a more severe case, Gemini provides an option with zero overlap when asked about restaurants open between 22:30 and 23:45, without acknowledging the limitation:

\begin{samplebox}
\textbf{Gemini 2.5 Flash} -- \emph{Restaurant opening hours}

\vspace{0.25em}

Sitar Tandoori is open until 22:30 on Fridays. Would you like to book a table there?
\end{samplebox}

\subsection{Constraint drifts}
\label{sec:constraint_drifts}

Our analysis also finds a tendency of LLMs to relax, misinterpret, or even fabricate constraints in the dialogue setup. In the following example, the LLM was asked to provide restaurants that are open \emph{at least for an hour} during a specific time interval. However, this is changed to a more relaxed constraint (\emph{at least for a moment}) in  the response:

\begin{samplebox}
\textbf{Qwen3 235B A22B} -- \emph{Restaurant opening hours}

\vspace{0.25em}

Let me check the opening hours for these restaurants on Monday between 06:30 and 09:00.\\

After reviewing the list:\\

- Maharajah Tandoori Restaurant is open on Monday from 09:00 to 18:00. This means it opens exactly at 09:00, so it is open for a brief moment at the end of the specified window (06:30–09:00), starting right at 09:00.\\

None of the other listed restaurants have Monday hours that fall within or overlap the 06:30 to 09:00 timeframe.\\

So, the only restaurant open on Monday for at least a moment between 06:30 and 09:00 is:
{[}...{]}
\end{samplebox}

An example of a fabricated constraint is a price range filter that Qwen3 30B invents to select only a subset of the valid restaurants open in the given time interval:

\begin{samplebox}
\textbf{Qwen3 30B A3B} -- \emph{Restaurant opening hours}

\vspace{0.25em}

Here are the moderate-priced Chinese restaurants open on Monday between 15:45 and 00:00: Jinling Noodle Bar and Shanghai Family Restaurant. Would you like to make a reservation at any of these?
\end{samplebox}

\section{Ablations}
\label{sec:appendix_ablations}

\subsection{Detailed results}
\label{sec:appendix_ablations_results}

\begin{figure*}[htbp]
    \centering
    \includegraphics[width=\textwidth]{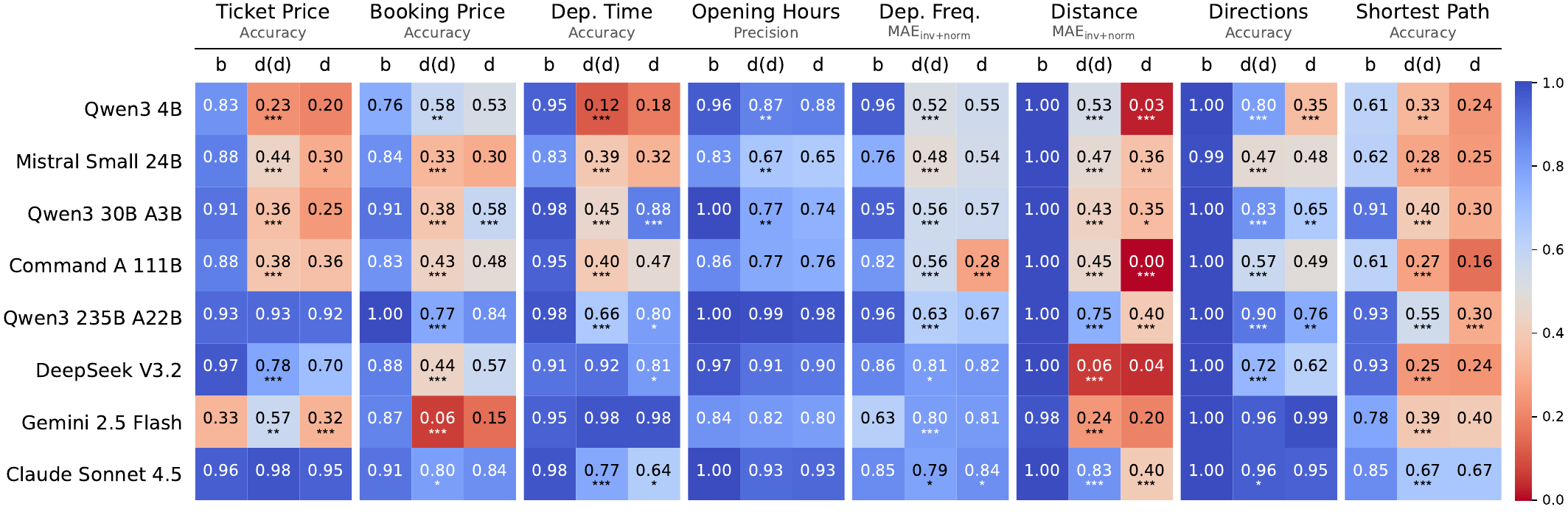}
    \caption{Detailed results for the \emph{Dialogue with reduced domains} (d(d)) ablation, compared with the \emph{Baseline} (b) and \emph{Dialogue} (d) setups. Asterisks indicate significant differences between setups in neighboring columns (t-test, $*\colon p < 0.05$, $**\colon p < 0.01$, ${*}{*}{*}\colon p < 0.001$).}
    \label{fig:results_ablation_dialogue_domains_full}
\end{figure*}

\begin{figure*}[htbp]
    \centering
    \includegraphics[width=\textwidth]{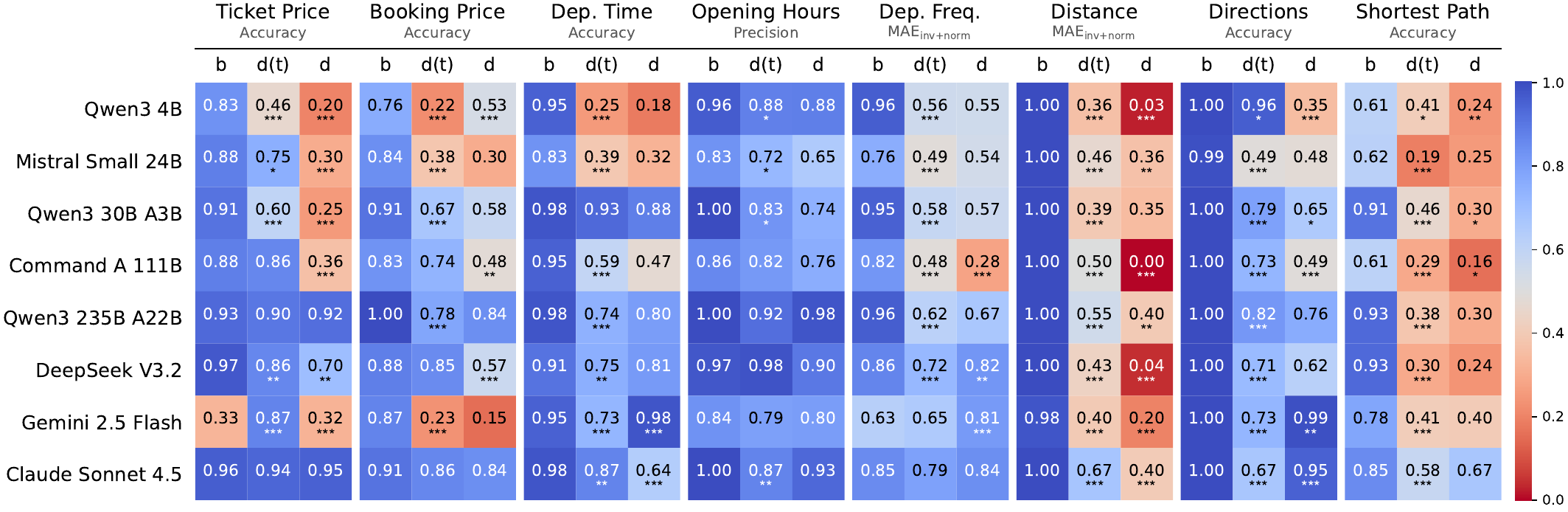}
    \caption{Detailed results for the \emph{Dialogue without tools} (d(t)) ablation, compared with the \emph{Baseline} (b) and \emph{Dialogue} (d) setups. Asterisks indicate significant differences between setups in neighboring columns (t-test, $*\colon p < 0.05$, $**\colon p < 0.01$, ${*}{*}{*}\colon p < 0.001$).}
    \label{fig:results_ablation_dialogue_no_tools_full}
\end{figure*}

\begin{figure*}[htbp]
    \centering
    \includegraphics[width=\textwidth]{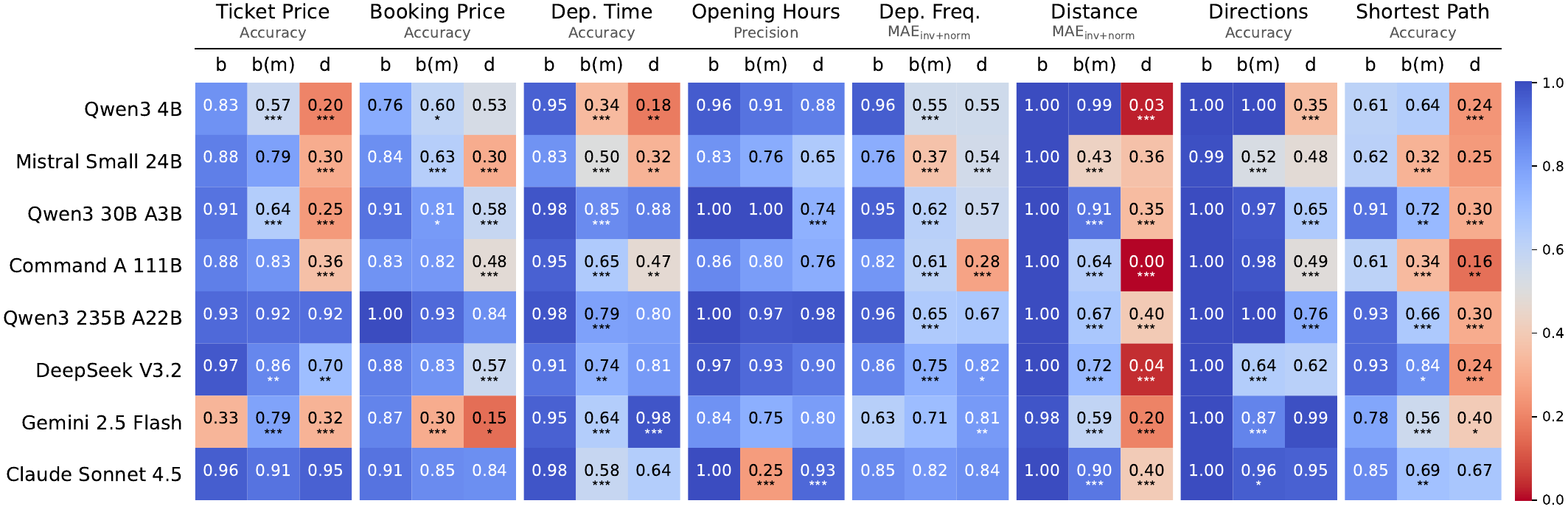}
    \caption{Detailed results for the \emph{Multi-turn baseline} (b(m)) ablation, compared with the \emph{Baseline} (b) and \emph{Dialogue} (d) setups. Asterisks indicate significant differences between setups in neighboring columns (t-test, $*\colon p < 0.05$, $**\colon p < 0.01$, ${*}{*}{*}\colon p < 0.001$).}
    \label{fig:results_ablation_baseline_multi_full}
\end{figure*}

\begin{figure*}[htbp]
    \centering
    \includegraphics[width=\textwidth]{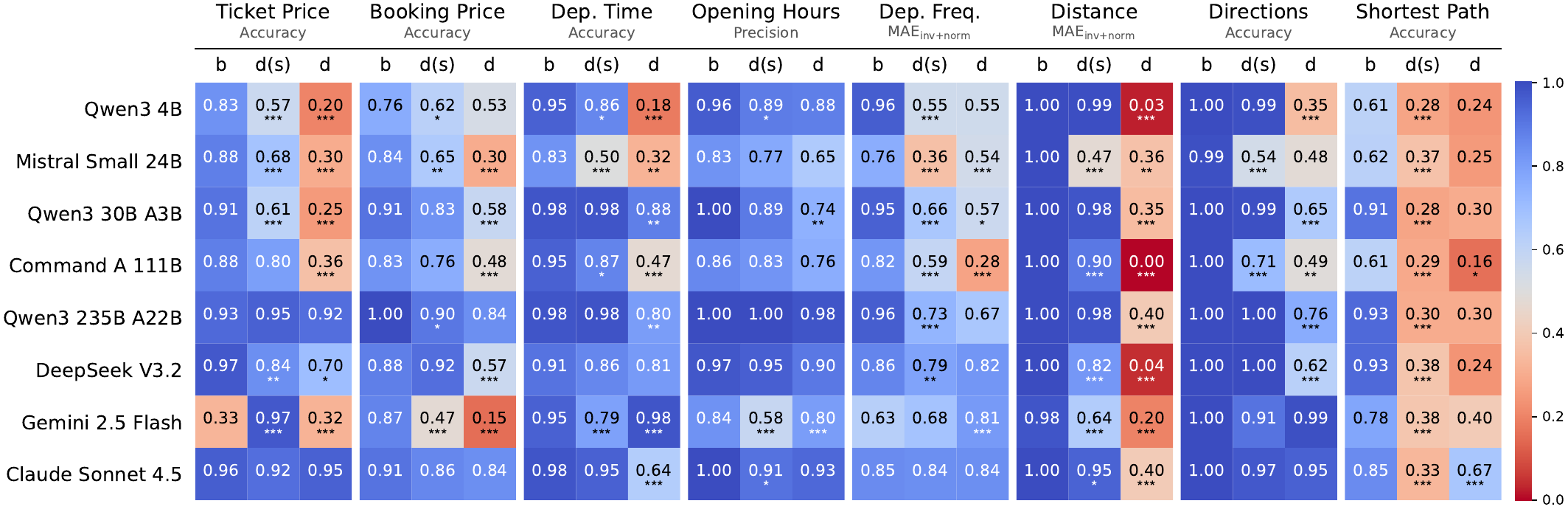}
    \caption{Detailed results for the \emph{Single-turn dialogue} (d(s)) ablation, compared with the \emph{Baseline} (b) and \emph{Dialogue} (d) setups. Asterisks indicate significant differences between setups in neighboring columns (t-test, $*\colon p < 0.05$, $**\colon p < 0.01$, ${*}{*}{*}\colon p < 0.001$).}
    \label{fig:results_ablation_dialogue_single_full}
\end{figure*}

\begin{figure*}[htbp]
    \centering
    \includegraphics[width=\textwidth]{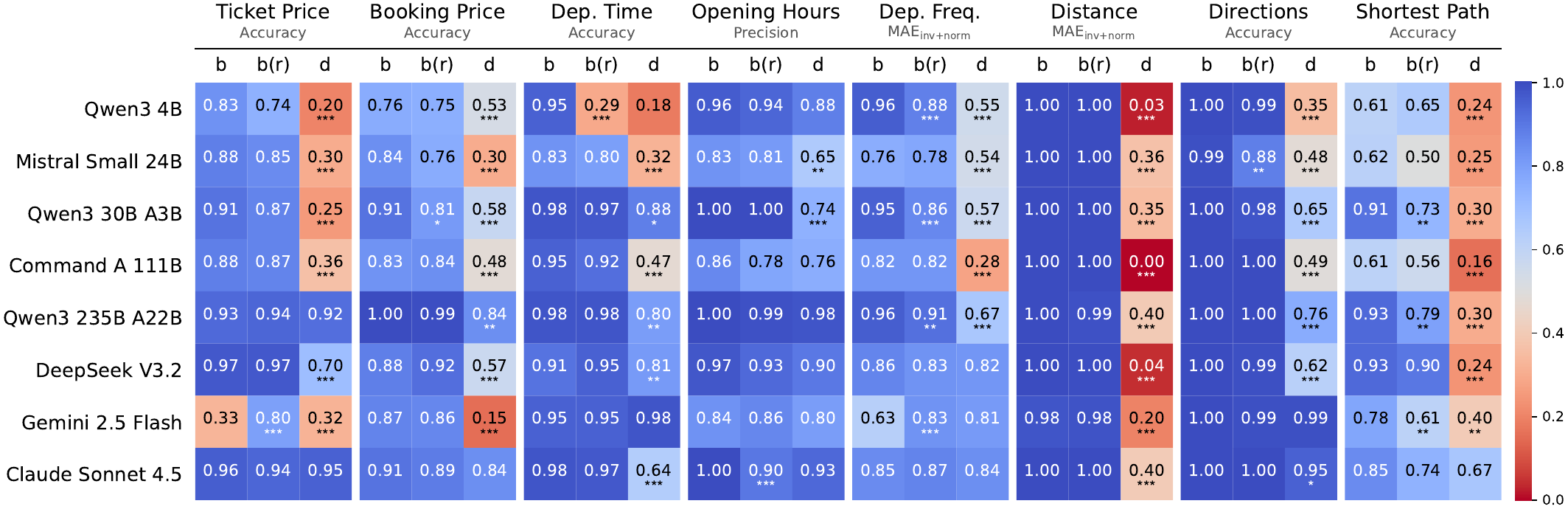}
    \caption{Detailed results for the \emph{Baseline with dialogue role} (b(r)) ablation, compared with the \emph{Baseline} (b) and \emph{Dialogue} (d) setups. Asterisks indicate significant differences between setups in neighboring columns (t-test, $*\colon p < 0.05$, $**\colon p < 0.01$, ${*}{*}{*}\colon p < 0.001$).}
    \label{fig:results_ablation_role_full}
\end{figure*}

\begin{figure*}[htbp]
    \centering
    \includegraphics[width=\textwidth]{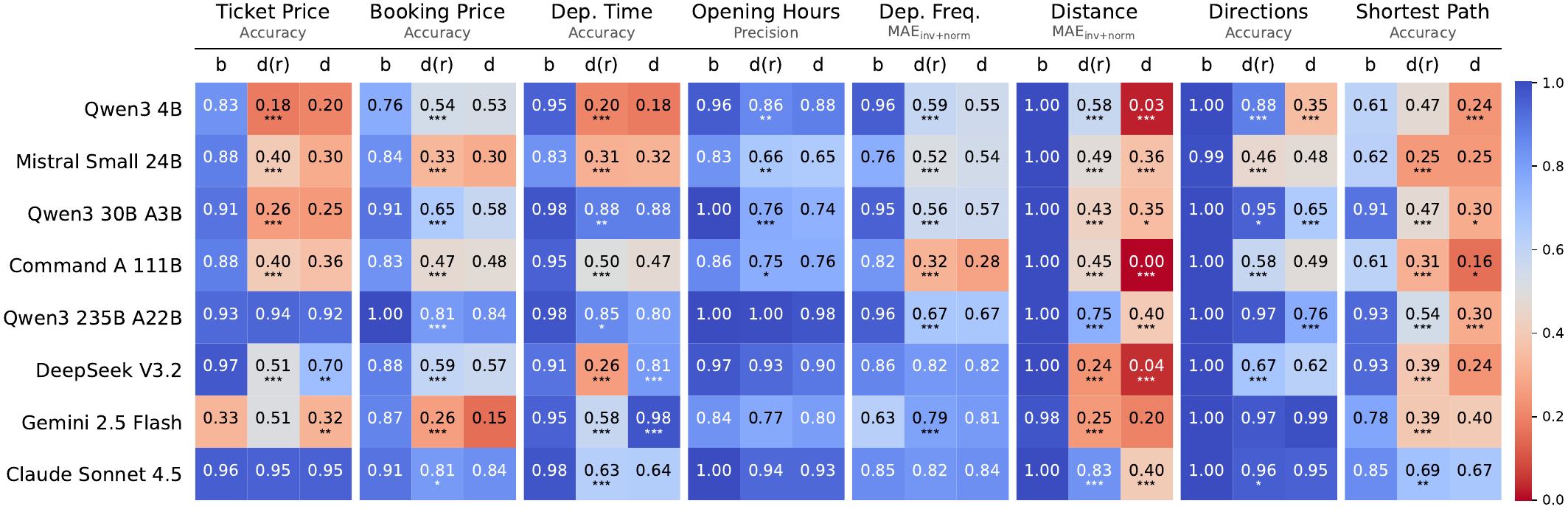}
    \caption{Detailed results for the \emph{Dialogue with reasoning instructions} (d(r)) ablation, compared with the \emph{Baseline} (b) and \emph{Dialogue} (d) setups. Asterisks indicate significant differences between setups in neighboring columns (t-test, $*\colon p < 0.05$, $**\colon p < 0.01$, ${*}{*}{*}\colon p < 0.001$).}
    \label{fig:results_ablation_dialogue_reasoning_full}
\end{figure*}

Figures~\ref{fig:results_ablation_dialogue_domains_full}--\ref{fig:results_ablation_dialogue_reasoning_full} show the detailed ablation results by task and model.

\subsection{Experiments}
\label{sec:appendix_ablations_exp}

\paragraph{Dialogue with reduced domains} As illustrated in Figure \ref{fig:ablation_dialogue_reduced_domains}, this ablation removes half of the domains and the corresponding tools, in particular those that are not used for a given task. The conversation template remains unchanged.

\paragraph{Dialogue without tools} Figure \ref{fig:ablation_dialogue_no_tools} shows the modifications made to the dialogue setup to create the variant without tools. The changes are as follows:
\begin{enumerate}
    \item Tool calls are removed from conversation histories and structured data retrieved by these tools are moved to the system prompt
    \item Instructions about tool use are replaced with instructions on how to use the provided context data
    \item Tool schemas are \emph{not} passed to LLMs during inference
\end{enumerate}

\paragraph{Single-turn dialogue} To transform the original dialogue setup to a single-turn variant, we merge all user's queries in a conversation history into a single query. For example, if the user first asks about trains with specific criteria and then asks about departure frequency in another turn, these questions are merged into a single user message that provides all information. The modifications are illustrated in Figure \ref{fig:ablation_dialogue_single_turn}.

\paragraph{Multi-turn baseline} Specific changes made to transform the baseline setup to a multi-turn setup are illustrated in Figure \ref{fig:ablation_baseline_multi}. In particular, we make the following changes:
\begin{enumerate}
    \item Message history with a single user prompt is extended with conversation turns from the dialogue setup (excluding tool calls)
    \item Role of the original \texttt{user} prompt is changed to \texttt{system}
    \item Query is moved to the last \texttt{user} message and replaced with a generic instruction: \emph{Given the \{domain\} data in JSON format below, answer the user's questions.}
\end{enumerate}

\paragraph{Baseline with dialogue role} As illustrated in Figure~\ref{fig:ablation_baseline_role}, we add two instructions related to the travel assistant role from the dialogue setup to baseline prompts:

\begin{itemize}
    \item \emph{You are a helpful assistant specialized in providing travel guidance for Cambridge.}
    \item \emph{Always maintain a friendly, polite, and helpful tone throughout the conversation.}
\end{itemize}

\paragraph{Dialogue with reasoning instructions} The changes for this variant involve an added instruction to use more time to generate if the response requires reasoning, as shown in Figure \ref{fig:ablation_dialogue_reasoning}.

\newpage
\begin{figure*}[h]
\centering
\small
\begin{tcolorbox}[
    colback=white,
    colframe=bordercolor,
    arc=3mm,
    boxrule=0.75pt,
    width=0.95\textwidth,
    left=5pt,
    right=5pt,
    top=5pt,
    bottom=5pt,
]
\ttfamily
\color{textgray}
\setlength{\parindent}{0pt}
\setlength{\parfillskip}{0pt plus 1fil}
\emergencystretch=3em
\sloppy

\begin{verbatim}
[
    {
        "role": "assistant",
        "content": "Hi, how can I help you today?"
    },
    {
        "role": "user",
        "content": "I'm looking for {{ food }} restaurants in the {{ area_adj }} part of town."
    },
    {
        "role": "placeholder",
        "content": "tool_call_restaurant"
    },
    {
        "role": "placeholder",
        "content": "tool_response_restaurant"
    },
    {
        "role": "assistant",
        "content": "I found {{ num_restaurants }} {{ food }} restaurants in the {{ area }}.
         Do you have any other preferences to help narrow this down?"
    },
    {
        "role": "user",
        "content": "Which of them are open on {{ weekday }} {{ time_specification }}?"
    }
]
\end{verbatim}

\end{tcolorbox}
\caption{Example conversation template in JSON format for the \emph{opening hours} task. During test example generation, placeholders in curly brackets are replaced with actual values and placeholder messages are filled with specific tool calls and tool responses.}
\label{fig:conversation_template}
\end{figure*}

\newpage
\begin{figure*}[h]
\centering
\small
\begin{tcolorbox}[
    colback=white,
    colframe=bordercolor,
    arc=3mm,
    boxrule=0.75pt,
    width=0.95\textwidth,
    left=5pt,
    right=5pt,
    top=5pt,
    bottom=5pt,
]
\ttfamily
\color{textgray}
\setlength{\parindent}{0pt}
\setlength{\parfillskip}{0pt plus 1fil}
\emergencystretch=3em
\sloppy

Given the following dialogue, generate a paraphrased version with the same meaning and structure. Paraphrase each user and assistant message in the dialogue using different words and phrasing. Keep the placeholders in curly braces in the same messages as in the original dialogue. All paraphrases should be friendly and conversational and not too formal. You must preserve all such placeholders from the original dialogue in their corresponding messages! Generate the dialogue as a JSON array of message objects.

\vspace{1.0em}
Previously generated paraphrased dialogues (generate something DIFFERENT from these, but keep the same meaning and structure):

\vspace{0.5em}
\{\{ previous\_paraphrases \}\}

\vspace{1.0em}
Original dialogue:

\vspace{0.5em}
\{\{ dialogue \}\}

\end{tcolorbox}
\caption{Prompt template used for generating paraphrases of conversation templates. Paraphrases generated in previous iterations are included in the prompt in subsequent inference calls to make the result more diverse.}
\label{fig:template_paraphrasing_prompt}
\end{figure*}

\newpage
\begin{figure*}[h]
\centering
\small
\begin{tcolorbox}[
    colback=white,
    colframe=bordercolor,
    arc=3mm,
    boxrule=0.75pt,
    width=0.95\textwidth,
    left=5pt,
    right=5pt,
    top=5pt,
    bottom=5pt,
]
\ttfamily
\color{textgray}
\setlength{\parindent}{0pt}
\setlength{\parfillskip}{0pt plus 1fil}
\emergencystretch=3em
\sloppy

You are a helpful assistant specialized in providing travel guidance for Cambridge. Your task is to help users find venues or transport based on their preferences, and make bookings and reservations. Based on the user's requirements, there are four different sub-tasks: restaurants, hotels, attractions, and trains.

\vspace{0.5em}
You can use the available tools to search for venues or transport, and make reservations or purchases. Based on the user's input, either respond with a message or call a tool.

\vspace{0.5em}
Start by greeting the user and asking what task they want to perform: "Hi, how can I help you today?"

\vspace{0.5em}
Always maintain a friendly, polite, and helpful tone throughout the conversation.

\vspace{0.5em}
Today is \{\{ weekday \}\} \{\{ date \}\} and the current time is \{\{ time \}\}.

\vspace{0.5em}
\# Task 1: Restaurant

\vspace{0.5em}
\#\# Task Description

\vspace{0.5em}
Help users find restaurants in Cambridge and/or make reservations.

\vspace{0.5em}
\#\# Task Instructions

\vspace{0.5em}
- Use the search\_restaurants tool to search for restaurants with specific parameters.

- Use the make\_restaurant\_reservation tool to make reservations for the user.

- After using the search\_restaurants tool, recommend matching restaurants to the user.

- If the search returns too many restaurants, ask for more constraints rather than offering to make a reservation.

\vspace{1.0em}
\# Task 2: Hotel

\vspace{0.5em}
\#\# Task Description

\vspace{0.5em}
Help users find hotels in Cambridge and/or make bookings.

\vspace{0.5em}
\#\# Task Instructions

\vspace{0.5em}
- Use the search\_hotels tool to search for hotels with specific parameters.

- Use the book\_hotel tool to book accommodation for the user.

- After using the search\_hotels tool, recommend matching hotels to the user.

- If the search returns too many hotels, ask for more constraints rather than offering to make a booking.

\vspace{1.0em}
\# Task 3: Attraction

\vspace{0.5em}
\#\# Task Description

\vspace{0.5em}
Help users find attractions in Cambridge.

\vspace{0.5em}
\#\# Task Instructions
\vspace{0.5em}

- Use the search\_attractions tool to search for attractions with specific parameters.

- After using the search\_attractions tool, recommend matching attractions to the user.

- If the search returns too many attractions, ask for more constraints.

\vspace{1.0em}
\# Task 4: Train

\vspace{0.5em}
\#\# Task Description

\vspace{0.5em}
Help users find train connections to/from Cambridge and/or buy tickets.

\vspace{0.5em}
\#\# Task Instructions

\vspace{0.5em}
- Use the search\_trains tool to search for trains with specific parameters.

- Use the buy\_train\_tickets tool to buy train tickets for the user.

- After using the search\_trains tool, recommend matching trains to the user.

- If the search returns too many trains, ask for more constraints rather than offering to purchase tickets.

- A return ticket costs exactly twice the price of a single ticket.

\end{tcolorbox}
\caption{Prompt template for the task-oriented dialogue system used in the \emph{Dialogue} setup in our experiments, adapted from \citep{xu-etal-2024-rethinking}.}
\label{fig:dialogue_prompt_template}
\end{figure*}

\newpage
\begin{figure*}[h]
\centering
\small
\begin{tcolorbox}[
    colback=white,
    colframe=bordercolor,
    arc=3mm,
    boxrule=0.75pt,
    width=0.95\textwidth,
    left=5pt,
    right=5pt,
    top=5pt,
    bottom=5pt,
]
\ttfamily
\color{textgray}
\setlength{\parindent}{0pt}
\setlength{\parfillskip}{0pt plus 1fil}
\emergencystretch=3em
\sloppy

You are a helpful assistant specialized in providing travel guidance for Cambridge. Your task is to help users find venues or transport based on their preferences, and make bookings and reservations. Based on the user's requirements, there are four different sub-tasks: restaurants, hotels, attractions, and trains.

\vspace{0.5em}
You can use the available tools to search for venues or transport, and make reservations or purchases. Based on the user's input, either respond with a message or call a tool.

\vspace{0.5em}
Start by greeting the user and asking what task they want to perform: "Hi, how can I help you today?"

\vspace{0.5em}
Always maintain a friendly, polite, and helpful tone throughout the conversation.

\vspace{0.5em}
Today is \{\{ weekday \}\} \{\{ date \}\} and the current time is \{\{ time \}\}.

\vspace{0.5em}
\# Task 1: Restaurant

\vspace{0.5em}
\#\# Task Description

\vspace{0.5em}
Help users find restaurants in Cambridge and/or make reservations.

\vspace{0.5em}
\#\# Task Instructions

\vspace{0.5em}
- Use the search\_restaurants tool to search for restaurants with specific parameters.

- Use the make\_restaurant\_reservation tool to make reservations for the user.

- After using the search\_restaurants tool, recommend matching restaurants to the user.

- If the search returns too many restaurants, ask for more constraints rather than offering to make a reservation.

\vspace{1.0em}
\# Task 2: Hotel

\vspace{0.5em}
\#\# Task Description

\vspace{0.5em}
Help users find hotels in Cambridge and/or make bookings.

\vspace{0.5em}
\#\# Task Instructions

\vspace{0.5em}
- Use the search\_hotels tool to search for hotels with specific parameters.

- Use the book\_hotel tool to book accommodation for the user.

- After using the search\_hotels tool, recommend matching hotels to the user.

- If the search returns too many hotels, ask for more constraints rather than offering to make a booking.

\vspace{1.0em}
\# Task 3: Attraction

\vspace{0.5em}
\#\# Task Description

\vspace{0.5em}
Help users find attractions in Cambridge.

\vspace{0.5em}
\#\# Task Instructions
\vspace{0.5em}

- Use the search\_attractions tool to search for attractions with specific parameters.

- After using the search\_attractions tool, recommend matching attractions to the user.

- If the search returns too many attractions, ask for more constraints.

\vspace{1.0em}
\# Task 4: Train

\vspace{0.5em}
\#\# Task Description

\vspace{0.5em}
Help users find train connections to/from Cambridge and/or buy tickets.

\vspace{0.5em}
\#\# Task Instructions

\vspace{0.5em}
- Use the search\_trains tool to search for trains with specific parameters.

- Use the buy\_train\_tickets tool to buy train tickets for the user.

- After using the search\_trains tool, recommend matching trains to the user.

- If the search returns too many trains, ask for more constraints rather than offering to purchase tickets.

- A return ticket costs exactly twice the price of a single ticket.

\vspace{0.5em}
\#\# General Instructions

\vspace{0.5em}
- Your responses should be concise and conversational

- Each response should consist of at most two short sentences.

\end{tcolorbox}
\caption{Prompt template for the task-oriented dialogue system used in the \emph{Dialogue-concise} setup in our experiments, adapted from \citep{xu-etal-2024-rethinking}. Note that it differs from the prompt for the \emph{Dialogue} setup (Figure~\ref{fig:dialogue_prompt_template}) only in the last section (``General Instructions'').}
\label{fig:dialogue_concise_prompt_template}
\end{figure*}

\begin{figure*}[h]
\centering
\small
\begin{tcolorbox}[
    colback=white,
    colframe=bordercolor,
    arc=3mm,
    boxrule=0.75pt,
    width=0.95\textwidth,
    left=5pt,
    right=5pt,
    top=5pt,
    bottom=5pt,
]
\ttfamily
\color{textgray}
\setlength{\parindent}{0pt}
\setlength{\parfillskip}{0pt plus 1fil}
\emergencystretch=3em
\sloppy

You are a helpful assistant specialized in providing travel guidance for Cambridge. Your task is to help users find venues or transport based on their preferences, and make bookings and reservations. Based on the user's requirements, there are four different sub-tasks: restaurants, hotels, attractions, and trains.

\vspace{0.5em}
You can use the available tools to search for venues or transport, and make reservations or purchases. Based on the user's input, either respond with a message or call a tool.

\vspace{0.5em}
Start by greeting the user and asking what task they want to perform: "Hi, how can I help you today?"

\vspace{0.5em}
Always maintain a friendly, polite, and helpful tone throughout the conversation.

\vspace{0.5em}
Today is \{\{ weekday \}\} \{\{ date \}\} and the current time is \{\{ time \}\}.

\vspace{0.5em}
\# Task 1: Restaurant

\vspace{0.5em}
\#\# Task Description

\vspace{0.5em}
Help users find restaurants in Cambridge and/or make reservations.

\vspace{0.5em}
\#\# Task Instructions

\vspace{0.5em}
- Use the search\_restaurants tool to search for restaurants with specific parameters.

- Use the make\_restaurant\_reservation tool to make reservations for the user.

- After using the search\_restaurants tool, recommend matching restaurants to the user.

- If the search returns too many restaurants, ask for more constraints rather than offering to make a reservation.

\vspace{1.0em}
\# Task 2: Hotel

\vspace{0.5em}
\#\# Task Description

\vspace{0.5em}
Help users find hotels in Cambridge and/or make bookings.

\vspace{0.5em}
\#\# Task Instructions

\vspace{0.5em}
- Use the search\_hotels tool to search for hotels with specific parameters.

- Use the book\_hotel tool to book accommodation for the user.

- After using the search\_hotels tool, recommend matching hotels to the user.

- If the search returns too many hotels, ask for more constraints rather than offering to make a booking.

\vspace{1.0em}
\removed{\# Task 3: Attraction}

\vspace{0.5em}
\removed{\#\# Task Description}

\vspace{0.5em}
\removed{Help users find attractions in Cambridge.}

\vspace{0.5em}
\removed{\#\# Task Instructions}
\vspace{0.5em}

\removed{- Use the search\_attractions tool to search for attractions with specific parameters.}

\removed{- After using the search\_attractions tool, recommend matching attractions to the user.}

\removed{- If the search returns too many attractions, ask for more constraints.}

\vspace{1.0em}
\removed{\# Task 4: Train}

\vspace{0.5em}
\removed{\#\# Task Description}

\vspace{0.5em}
\removed{Help users find train connections to/from Cambridge and/or buy tickets.}

\vspace{0.5em}
\removed{\#\# Task Instructions}

\vspace{0.5em}
\removed{- Use the search\_trains tool to search for trains with specific parameters.}

\removed{- Use the buy\_train\_tickets tool to buy train tickets for the user.}

\removed{- After using the search\_trains tool, recommend matching trains to the user.}

\removed{- If the search returns too many trains, ask for more constraints rather than offering to purchase tickets.}

\removed{- A return ticket costs exactly twice the price of a single ticket.}

\end{tcolorbox}
\caption{Changes made to the prompt for the \textit{Dialogue with reduced domains} ablation. Text highlighted in red indicates removed content. This example illustrates the changes made to the \emph{distance} and \emph{opening hours} tasks, where we remove the \texttt{attractions} and \texttt{trains} domains.}
\label{fig:ablation_dialogue_reduced_domains}
\end{figure*}

\begin{figure*}[h]
\centering
\small
\begin{tcolorbox}[
    colback=white,
    colframe=bordercolor,
    arc=3mm,
    boxrule=0.75pt,
    width=0.95\textwidth,
    left=5pt,
    right=5pt,
    top=5pt,
    bottom=5pt,
]
\ttfamily
\color{textgray}
\setlength{\parindent}{0pt}
\setlength{\parfillskip}{0pt plus 1fil}
\emergencystretch=3em
\sloppy

\textbf{SYSTEM:} You are a helpful assistant specialized in providing travel guidance for Cambridge. Your task is to help users find venues or transport based on their preferences, and make bookings and reservations. Based on the user's requirements, there are four different sub-tasks: restaurants, hotels, attractions, and trains.

\vspace{0.5em}
You can use the available tools to search for venues or transport, and make reservations or purchases.\removed{ Based on the user's input, either respond with a message or call a tool.}\added{ Based on the user's input, either answer their questions or confirm a booking, purchase or reservation. We will parse your confirmation and make the actual booking, purchase or reservation later.}

\vspace{0.5em}
Start by greeting the user and asking what task they want to perform: "Hi, how can I help you today?"

\vspace{0.5em}
Always maintain a friendly, polite, and helpful tone throughout the conversation.

\vspace{0.5em}
Today is \{\{ weekday \}\} \{\{ date \}\} and the current time is \{\{ time \}\}.

\vspace{0.5em}
\# Task 1: Trains

\vspace{0.5em}
\added{\#\# Data}

\vspace{0.5em}
\added{[\{``trainID'': ``TR0393'', ``departure'': ``broxbourne'', ``destination'': ``cambridge'', ``day'': ``thursday'', ``leaveAt'': ``06:07'', ``arriveBy'': ``07:07'', ``duration'': ``60 minutes'', ``price\_standard'': ``18.40 pounds'', ``price\_first\_class'': ``27.60 pounds''\}, \{``trainID'': ``TR13021'', ``departure'': ``broxbourne'', ``destination'': ``cambridge'', ``day'': ``thursday'', ``leaveAt'': ``06:21'', ``arriveBy'': ``07:21'', ``duration'': ``60 minutes'', ``price\_standard'': ``18.40 pounds'', ``price\_first\_class'': ``27.60 pounds''\}, ...]}

\vspace{0.5em}
\#\# Task Description

\vspace{0.5em}
Help users find train connections to/from Cambridge and/or buy tickets.

\vspace{0.5em}
\#\# Task Instructions

\vspace{0.5em}
\removed{- Use the search\_trains tool to search for trains with specific parameters.}

\added{- Use the train data to search for trains with specific parameters.}

\removed{- Use the buy\_train\_tickets tool to buy train tickets for the user.}

\added{- To buy tickets for the user, just say that you are making the purchase. We will parse it and do the actual booking later.}

\removed{- After using the search\_trains tool, recommend matching trains to the user.}

\added{- After searching in the train data, recommend matching trains to the user.}

\removed{- If the search returns too many trains, ask for more constraints rather than offering to purchase tickets.}

\added{- If you find too many trains, ask for more constraints rather than offering to purchase tickets.}

- A return ticket costs exactly twice the price of a single ticket.

\vspace{0.5em}
\removed{[ \ldots TOOL SCHEMAS \ldots ]}

\vspace{1.0em}
\textbf{ASSISTANT:} Hi, how can I help you today?

\textbf{USER:} I'm looking for trains from Broxbourne to Cambridge on Thursday.

\removed{\textbf{ASSISTANT:} \{``type'': ``function'', ``id'': ``205d4e'', ``function'': \{``name'': ``search\_trains'', ``arguments'': \{``departure'': ``broxbourne'', ``destination'': ``cambridge'', ``weekday'': ``thursday''\}\}}

\removed{\textbf{TOOL:} [\{``trainID'': ``TR0393'', ``departure'': ``broxbourne'', ``destination'': ``cambridge'', ``day'': ``thursday'', ``leaveAt'': ``06:07'', ``arriveBy'': ``07:07'', ``duration'': ``60 minutes'', ``price\_standard'': ``18.40 pounds'', ``price\_first\_class'': ``27.60 pounds''\}, ...]}

\textbf{ASSISTANT:} I found 59 trains leaving on Thursday. Do you have a preferred departure time?

\textbf{USER:} How often on average do trains run between 20:00 and midnight?

\end{tcolorbox}
\caption{Changes made to the prompt and the message history for the \emph{Dialogue without tools} ablation. Text highlighted in red indicates removed content, while a green highlight indicates added content. The ablation consists of the following modifications: tool schemas are removed, the data from tool responses are added to the system prompt, the instructions that refer to tools are modified to refer to the data.}
\label{fig:ablation_dialogue_no_tools}
\end{figure*}

\begin{figure*}[h]
\centering
\small
\begin{tcolorbox}[
    colback=white,
    colframe=bordercolor,
    arc=3mm,
    boxrule=0.75pt,
    width=0.95\textwidth,
    left=5pt,
    right=5pt,
    top=5pt,
    bottom=5pt,
]
\ttfamily
\color{textgray}
\setlength{\parindent}{0pt}
\setlength{\parfillskip}{0pt plus 1fil}
\emergencystretch=3em
\sloppy

\textbf{SYSTEM:} You are a helpful assistant specialized in providing travel guidance for Cambridge. Your task is to help users find venues or transport based on their preferences, and make bookings and reservations. Based on the user's requirements, there are four different sub-tasks: restaurants, hotels, attractions, and trains.
\vspace{0.5em}

You can use the available tools to search for venues or transport, and make reservations or purchases.\removed{ Based on the user's input, either respond with a message or call a tool.}\added{ Based on the user's input, either answer their questions or confirm a booking, purchase or reservation. We will parse your confirmation and make the actual booking, purchase or reservation later.}

\vspace{0.5em}
Start by greeting the user and asking what task they want to perform: "Hi, how can I help you today?"

\vspace{0.5em}
Always maintain a friendly, polite, and helpful tone throughout the conversation.

\vspace{0.5em}
Today is \{\{ weekday \}\} \{\{ date \}\} and the current time is \{\{ time \}\}.

\vspace{0.5em}
\# Task 1: Trains

\vspace{0.5em}
\added{\#\# Data}

\vspace{0.5em}
\added{[\{``trainID'': ``TR0393'', ``departure'': ``broxbourne'', ``destination'': ``cambridge'', ``day'': ``thursday'', ``leaveAt'': ``06:07'', ``arriveBy'': ``07:07'', ``duration'': ``60 minutes'', ``price\_standard'': ``18.40 pounds'', ``price\_first\_class'': ``27.60 pounds''\}, \{``trainID'': ``TR13021'', ``departure'': ``broxbourne'', ``destination'': ``cambridge'', ``day'': ``thursday'', ``leaveAt'': ``06:21'', ``arriveBy'': ``07:21'', ``duration'': ``60 minutes'', ``price\_standard'': ``18.40 pounds'', ``price\_first\_class'': ``27.60 pounds''\}, ...]}

\vspace{0.5em}
\#\# Task Description

\vspace{0.5em}
Help users find train connections to/from Cambridge and/or buy tickets.

\vspace{0.5em}
\#\# Task Instructions

\vspace{0.5em}
\removed{- Use the search\_trains tool to search for trains with specific parameters.}

\added{- Use the train data to search for trains with specific parameters.}

\removed{- Use the buy\_train\_tickets tool to buy train tickets for the user.}

\added{- To buy tickets for the user, just say that you are making the purchase. We will parse it and do the actual booking later.}

\removed{- After using the search\_trains tool, recommend matching trains to the user.}

\added{- After searching in the train data, recommend matching trains to the user.}

\removed{- If the search returns too many trains, ask for more constraints rather than offering to purchase tickets.}

\added{- If you find too many trains, ask for more constraints rather than offering to purchase tickets.}

- A return ticket costs exactly twice the price of a single ticket.

\vspace{0.5em}
\removed{[ \ldots TOOL SCHEMAS \ldots ]}

\vspace{1.0em}
\removed{\textbf{ASSISTANT:} Hi, how can I help you today?}

\removed{\textbf{USER:} I'm looking for trains from Broxbourne to Cambridge on Thursday.}

\removed{\textbf{ASSISTANT:} \{``type'': ``function'', ``id'': ``205d4e'', ``function'': \{``name'': ``search\_trains'', ``arguments'': \{``departure'': ``broxbourne'', ``destination'': ``cambridge'', ``weekday'': ``thursday''\}\}}

\removed{\textbf{TOOL:} [\{``trainID'': ``TR0393'', ``departure'': ``broxbourne'', ``destination'': ``cambridge'', ``day'': ``thursday'', ``leaveAt'': ``06:07'', ``arriveBy'': ``07:07'', ``duration'': ``60 minutes'', ``price\_standard'': ``18.40 pounds'', ``price\_first\_class'': ``27.60 pounds''\}, ...]}

\removed{\textbf{ASSISTANT:} I found 59 trains leaving on Thursday. Do you have a preferred departure time?}

\removed{\textbf{USER:} I'm looking for trains from Broxbourne to Cambridge on Thursday. How often on average do trains run between 20:00 and midnight?}

\added{\textbf{USER:} I'm looking for trains from Broxbourne to Cambridge on Thursday. How often on average do they run between 20:00 and midnight?}

\end{tcolorbox}
\caption{Changes made to the prompt and the message history for the \emph{Single-turn dialogue} ablation. Text highlighted in red indicates removed content, while a green highlight indicates added content. The ablation consists of the following modifications: all \texttt{user} and \texttt{assistant} messages are removed from the message history and replaced by a single \texttt{user} message that combines all previous \texttt{user} queries (if they are applicable); other modifications are the same as in \emph{Dialogue without tools}.}
\label{fig:ablation_dialogue_single_turn}
\end{figure*}

\newpage
\begin{figure*}[h]
\centering
\small
\begin{tcolorbox}[
    colback=white,
    colframe=bordercolor,
    arc=3mm,
    boxrule=0.75pt,
    width=0.95\textwidth,
    left=5pt,
    right=5pt,
    top=5pt,
    bottom=5pt,
]
\ttfamily
\color{textgray}
\setlength{\parindent}{0pt}
\setlength{\parfillskip}{0pt plus 1fil}
\emergencystretch=3em
\sloppy

\removed{\textbf{USER:}}\added{\textbf{SYSTEM:}} Given the list of trains in JSON format below, \removed{calculate how often on average do trains run between 20:00 and midnight.}\added{answer the user's questions.}

\vspace{0.5em}
Trains:

[\{``trainID'': ``TR0393'', ``departure'': ``broxbourne'', ``destination'': ``cambridge'', ``day'': ``thursday'', ``leaveAt'': ``06:07'', ``arriveBy'': ``07:07'', ``duration'': ``60 minutes'', ``price\_standard'': ``18.40 pounds'', ``price\_first\_class'': ``27.60 pounds''\}, ...]

\vspace{0.5em}
\added{\textbf{ASSISTANT:} Hi, how can I help you today?}

\added{\textbf{USER:} I'm looking for trains from Broxbourne to Cambridge on Thursday.}

\added{\textbf{ASSISTANT:} I found 59 trains leaving on Thursday. Do you have a preferred departure time?}

\added{\textbf{USER:} How often on average do trains run between 20:00 and midnight?}

\end{tcolorbox}
\caption{Changes made to the prompt and the message history for the \emph{Multi-turn baseline} ablation. Text highlighted in red indicates removed content, while a green highlight indicates added content. The ablation consists of the following modifications: the role of the original prompt is changed to \texttt{system}, the turns from the dialogue setup are added to the message history, and the query is moved from the original prompt to the last \texttt{user} message.}
\label{fig:ablation_baseline_multi}
\end{figure*}

\begin{figure*}[h]
\centering
\small
\begin{tcolorbox}[
    colback=white,
    colframe=bordercolor,
    arc=3mm,
    boxrule=0.75pt,
    width=0.95\textwidth,
    left=5pt,
    right=5pt,
    top=5pt,
    bottom=5pt,
]
\ttfamily
\color{textgray}
\setlength{\parindent}{0pt}
\setlength{\parfillskip}{0pt plus 1fil}
\emergencystretch=3em
\sloppy

\textbf{USER:} \added{You are a helpful assistant specialized in providing travel guidance for Cambridge.}

\added{Always maintain a friendly, polite, and helpful tone throughout the conversation.}

Given the list of trains in JSON format below, calculate how often on average do trains run between 20:00 and midnight.

\vspace{0.5em}
Trains:

[\{``trainID'': ``TR0393'', ``departure'': ``broxbourne'', ``destination'': ``cambridge'', ``day'': ``thursday'', ``leaveAt'': ``06:07'', ``arriveBy'': ``07:07'', ``duration'': ``60 minutes'', ``price\_standard'': ``18.40 pounds'', ``price\_first\_class'': ``27.60 pounds''\}, ...]

\end{tcolorbox}
\caption{Changes made to the prompt and the message history for the \emph{Baseline with dialogue role} ablation. Text highlighted in green indicates added content. Instructions related to the role of the task-oriented dialogue assistant are added to the baseline prompt.}
\label{fig:ablation_baseline_role}
\end{figure*}

\begin{figure*}[h]
\centering
\small
\begin{tcolorbox}[
    colback=white,
    colframe=bordercolor,
    arc=3mm,
    boxrule=0.75pt,
    width=0.95\textwidth,
    left=5pt,
    right=5pt,
    top=5pt,
    bottom=5pt,
]
\ttfamily
\color{textgray}
\setlength{\parindent}{0pt}
\setlength{\parfillskip}{0pt plus 1fil}
\emergencystretch=3em
\sloppy

\textbf{SYSTEM:} You are a helpful assistant specialized in providing travel guidance for Cambridge. Your task is to help users find venues or transport based on their preferences, and make bookings and reservations. Based on the user's requirements, there are four different sub-tasks: restaurants, hotels, attractions, and trains.

\vspace{0.5em}
You can use the available tools to search for venues or transport, and make reservations or purchases. Based on the user's input, either respond with a message or call a tool. \added{When the answer requires reasoning, take more time to think about the answer.}

\vspace{0.5em}
Start by greeting the user and asking what task they want to perform: "Hi, how can I help you today?"

\vspace{0.5em}
Always maintain a friendly, polite, and helpful tone throughout the conversation.

\vspace{0.5em}
Today is \{\{ weekday \}\} \{\{ date \}\} and the current time is \{\{ time \}\}.

\vspace{0.5em}
\# Task 1: Restaurant

\vspace{0.5em}
\#\# Task Description

\vspace{0.5em}
Help users find restaurants in Cambridge and/or make reservations.

\vspace{0.5em}
\#\# Task Instructions

- Use the search\_restaurants tool to search for restaurants with specific parameters.

- Use the make\_restaurant\_reservation tool to make reservations for the user.

- After using the search\_restaurants tool, recommend matching restaurants to the user.

- If the search returns too many restaurants, ask for more constraints rather than offering to make a reservation.

\vspace{1.0em}
[ \ldots MORE DOMAINS \ldots ]

\end{tcolorbox}
\caption{Changes made to the prompt and the message history for the \emph{Reasoning instructions} ablation. Text highlighted in green indicates added content. The prompt is the same as in the \emph{Dialogue} setup, with the added instruction for the model to use more time for reasoning when needed.}
\label{fig:ablation_dialogue_reasoning}
\end{figure*}

\newpage
\begin{figure*}[h]
\centering
\small
\begin{tcolorbox}[
    colback=white,
    colframe=bordercolor,
    arc=3mm,
    boxrule=0.75pt,
    width=0.95\textwidth,
    left=5pt,
    right=5pt,
    top=5pt,
    bottom=5pt,
]
\ttfamily
\color{textgray}
\setlength{\parindent}{0pt}
\setlength{\parfillskip}{0pt plus 1fil}
\emergencystretch=3em
\sloppy

You are given a response of a chat assistant that provides information about train ticket prices. Extract the total price that the assistant provided in the response. Only extract the value if it is explicitly mentioned as a total price in the response. Otherwise extract \textasciigrave null\textasciigrave. Parse the total price as a float number. Generate only the parsed answer without any additional text in the JSON format: \{"total\_price": <float>\}.

\vspace{0.8em}

Assistant's response:

\vspace{0.2em}

\{answer\}

\end{tcolorbox}
\caption{{Prompt template for the amounts parser used to extract values in the \emph{Train ticket price} and \emph{Hotel booking price} tasks. The answer is extracted from model output as a floating point number, representing the total amount for requested train tickets or hotel booking.}}
\label{fig:parser_amount_prompt}
\end{figure*}

\begin{figure*}[h]
\centering
\small
\begin{tcolorbox}[
    colback=white,
    colframe=bordercolor,
    arc=3mm,
    boxrule=0.75pt,
    width=0.95\textwidth,
    left=5pt,
    right=5pt,
    top=5pt,
    bottom=5pt,
]
\ttfamily
\color{textgray}
\setlength{\parindent}{0pt}
\setlength{\parfillskip}{0pt plus 1fil}
\emergencystretch=3em
\sloppy

You are given a response of a chat assistant that provides information about train departures. Extract the departure time that the assistant provided in the response as the latest departure time before sunset. Only extract the value if it is explicitly mentioned as a departure time in the response. Otherwise extract \textasciigrave null\textasciigrave. Parse the time in ISO format as HH:MM. Generate only the parsed answer without any additional text in the JSON format: \{"latest\_departure\_time": "HH:MM"\}

\vspace{0.8em}

Assistant's response:

\vspace{0.2em}

\{answer\}

\end{tcolorbox}
\caption{{Prompt template for the time parser used to extract values in the \emph{Train departure time} task. The answer is extracted from model output as a timestamp in ISO format (HH:MM), representing the departure time of the latest train that arrives to its destination before sunset.}}
\label{fig:parser_time_prompt}
\end{figure*}

\begin{figure*}[h]
\centering
\small
\begin{tcolorbox}[
    colback=white,
    colframe=bordercolor,
    arc=3mm,
    boxrule=0.75pt,
    width=0.95\textwidth,
    left=5pt,
    right=5pt,
    top=5pt,
    bottom=5pt,
]
\ttfamily
\color{textgray}
\setlength{\parindent}{0pt}
\setlength{\parfillskip}{0pt plus 1fil}
\emergencystretch=3em
\sloppy

You are given a response of a chat assistant that provides information about train departures. Extract the average interval (frequency) between train departures that the assistant provided in the response. This could be expressed in various ways such as 'every 2 hours', 'every 60 minutes', 'trains depart every hour', etc. Only extract the value if it is explicitly mentioned as a departure frequency or interval in the response. If the answer is a range, extract the average of the range. Otherwise extract \textasciigrave null\textasciigrave. Parse the interval in minutes as a float number. Generate only the parsed answer in the JSON format: \{"average\_interval\_minutes": <float>\}

\vspace{0.8em}

Assistant's response:

\vspace{0.2em}

\{answer\}

\end{tcolorbox}
\caption{{Prompt template for the time parser used to extract values in the \emph{Train departure frequency} task. The answer is extracted from model output as a floating point number representing the average time interval of departures in minutes.}}
\label{fig:parser_frequency_prompt}
\end{figure*}

\begin{figure*}[h]
\centering
\small
\begin{tcolorbox}[
    colback=white,
    colframe=bordercolor,
    arc=3mm,
    boxrule=0.75pt,
    width=0.95\textwidth,
    left=5pt,
    right=5pt,
    top=5pt,
    bottom=5pt,
]
\ttfamily
\color{textgray}
\setlength{\parindent}{0pt}
\setlength{\parfillskip}{0pt plus 1fil}
\emergencystretch=3em
\sloppy

You are given a response of a chat assistant that provides information about restaurants. Extract the names of the restaurants that the assistant identified as ones that meet the condition (i.e. open during the requested time) from the response. The restaurant names are given in the following list: \{restaurant\_names\}. Extract these names as a list of strings. Generate only the parsed answer without any additional text in the JSON format: \{"restaurant\_names": [...]\}.

\vspace{0.8em}

Assistant's response:

\vspace{0.2em}

\{answer\}

\end{tcolorbox}
\caption{{Prompt template for the restaurant names parser used to extract values in the \emph{Restaurant opening hours} task. The answer is extracted from model output as a list of restaurant names, representing the subset of restaurants that are open during the requested time interval.}}
\label{fig:parser_restaurants_prompt}
\end{figure*}

\begin{figure*}[h]
\centering
\small
\begin{tcolorbox}[
    colback=white,
    colframe=bordercolor,
    arc=3mm,
    boxrule=0.75pt,
    width=0.95\textwidth,
    left=5pt,
    right=5pt,
    top=5pt,
    bottom=5pt,
]
\ttfamily
\color{textgray}
\setlength{\parindent}{0pt}
\setlength{\parfillskip}{0pt plus 1fil}
\emergencystretch=3em
\sloppy

You are given a response of a chat assistant that provides information about hotels and restaurants. Extract the distance between the hotel and the restaurant that the assistant provided in the response. Only extract the value if it is explicitly mentioned as a distance in the response. Otherwise extract \textasciigrave null\textasciigrave. If a range is given, extract the average of the range. Parse the distance in meters as an integer number. Generate only the parsed answer without any additional text in the JSON format: \{"distance": <int>\}.

\vspace{0.8em}

Assistant's response:

\vspace{0.2em}

\{answer\}

\end{tcolorbox}
\caption{{Prompt template for the time parser used to extract values in the \emph{Distance between venues} task. The answer is extracted from model output as an integer, representing the distance in meters between specified venues.}}
\label{fig:parser_distance_prompt}
\end{figure*}

\begin{figure*}[h]
\centering
\small
\begin{tcolorbox}[
    colback=white,
    colframe=bordercolor,
    arc=3mm,
    boxrule=0.75pt,
    width=0.95\textwidth,
    left=5pt,
    right=5pt,
    top=5pt,
    bottom=5pt,
]
\ttfamily
\color{textgray}
\setlength{\parindent}{0pt}
\setlength{\parfillskip}{0pt plus 1fil}
\emergencystretch=3em
\sloppy

You are given a response of a chat assistant that provides information about directions. Extract whether the assistant answered in the response that {attraction\_name} is to the {asked\_direction} of {restaurant\_name}. If the assistant answers with just a yes or no without explanation, extract the answer. If the assistant does not provide the answer or says that they are in the same area, extract \textasciigrave unknown\textasciigrave. Generate only the parsed answer in the JSON format: \{"is\_\{asked\_direction\}": <yes/no/unknown>\}.

\vspace{0.8em}

Assistant's response:

\vspace{0.2em}

\{answer\}

\end{tcolorbox}
\caption{{Prompt template for the time parser used to extract values in the \emph{Directional relations} task. The answer is extracted from model output as one of three string values (yes/no/unknown), corresponding to a model's answer to a question about a venue being in a specific cardinal direction of another venue.}}
\label{fig:parser_directions_prompt}
\end{figure*}

\begin{figure*}[h]
\centering
\small
\begin{tcolorbox}[
    colback=white,
    colframe=bordercolor,
    arc=3mm,
    boxrule=0.75pt,
    width=0.95\textwidth,
    left=5pt,
    right=5pt,
    top=5pt,
    bottom=5pt,
]
\ttfamily
\color{textgray}
\setlength{\parindent}{0pt}
\setlength{\parfillskip}{0pt plus 1fil}
\emergencystretch=3em
\sloppy

You are given a response of a chat assistant that provides information about attractions. Extract the order to visit the attractions that the assistant provided in the response as the optimal order. The attraction names are given in the following list: \{attraction\_names\}.

Extract the names only if they are explicitly mentioned in the response. If the assistant does not provide the order, extract \textasciigrave null\textasciigrave. Generate the parsed answer in the JSON format: \{"order": [...]\}.

\vspace{0.8em}

Assistant's response:

\vspace{0.2em}

\{answer\}

\end{tcolorbox}
\caption{{Prompt template for the time parser used to extract values in the \emph{Shortest walking path} task. The answer is extracted from model output as one a list of attraction names representing the optimal order to visit the venues from a hotel.}}
\label{fig:parser_order_prompt}
\end{figure*}

\clearpage

\newpage

\newpage

\newpage
\begin{figure*}[htbp]
    \centering
    \includegraphics[width=\textwidth]{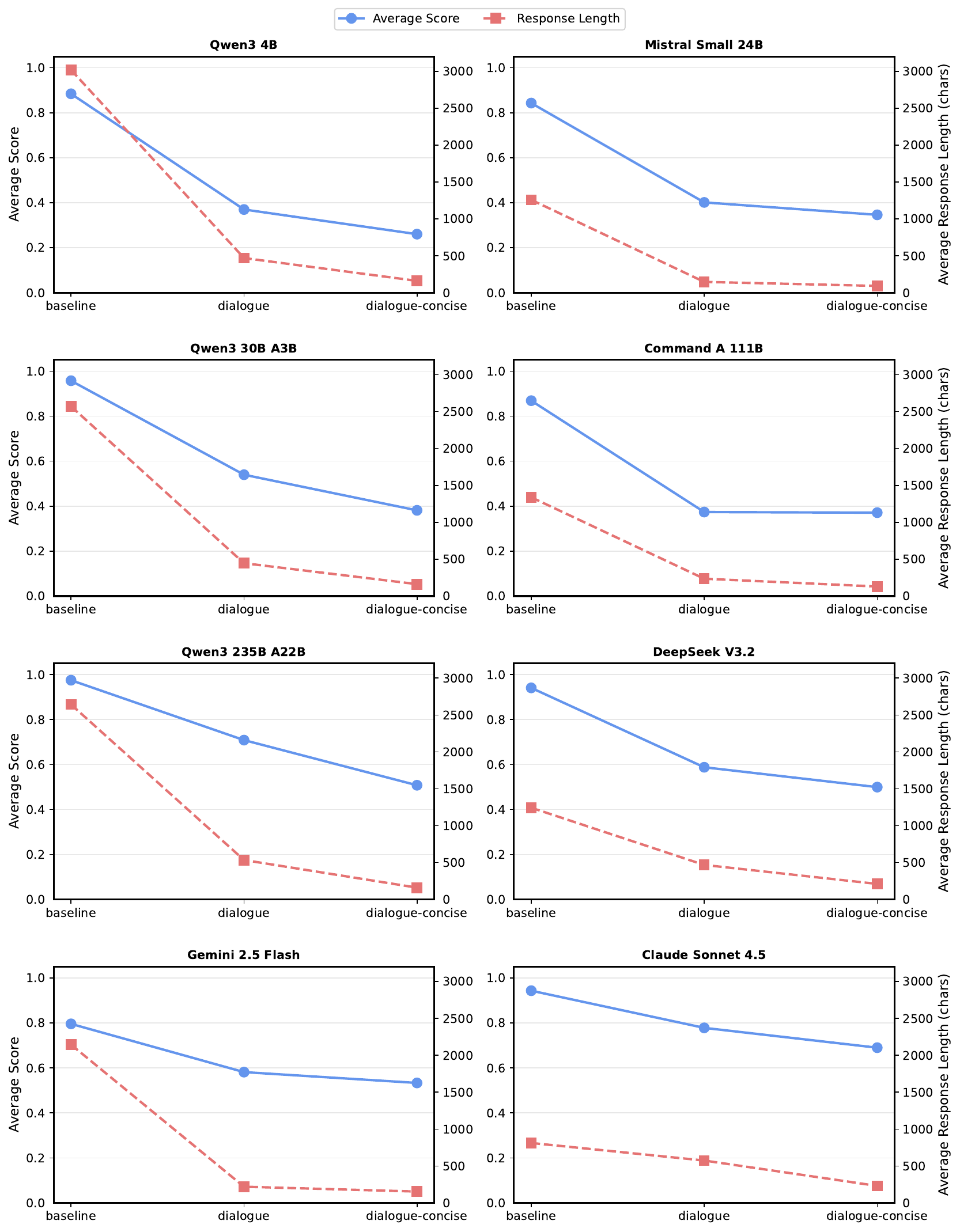}
    \caption{Scores and average response lengths in characters by LLM averaged over all tasks.}
    \label{fig:accuracy_vs_length}
\end{figure*}

\newpage
\begin{figure*}[htbp]
    \centering
    \includegraphics[width=\textwidth]{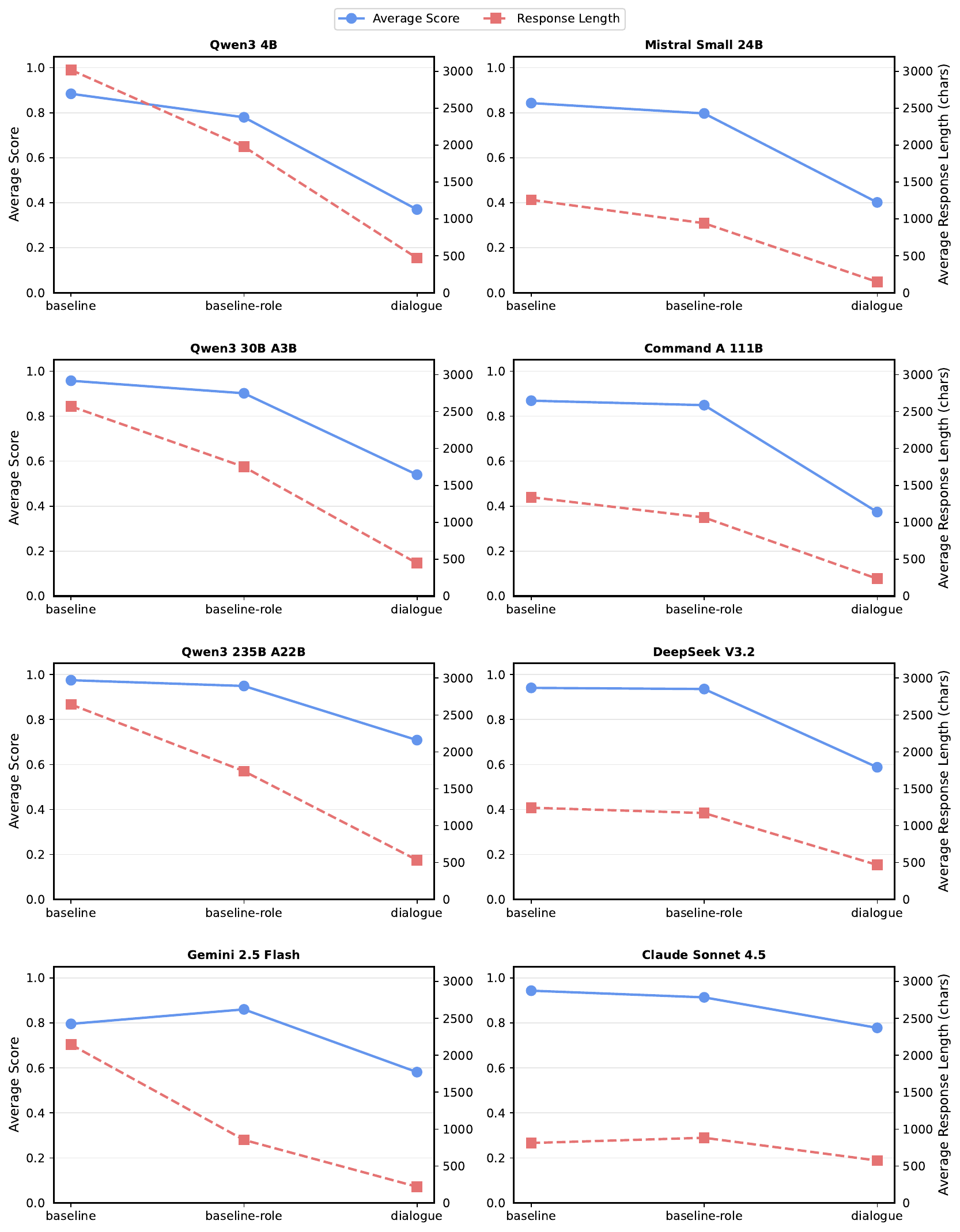}
    \caption{Comparison of scores and average response lengths in characters by LLM for the \textit{Baseline with dialogue role} ablation.}
    \label{fig:accuracy_vs_length_role}
\end{figure*}

\newpage
\begin{figure*}[htbp]
    \centering
    \includegraphics[width=\textwidth]{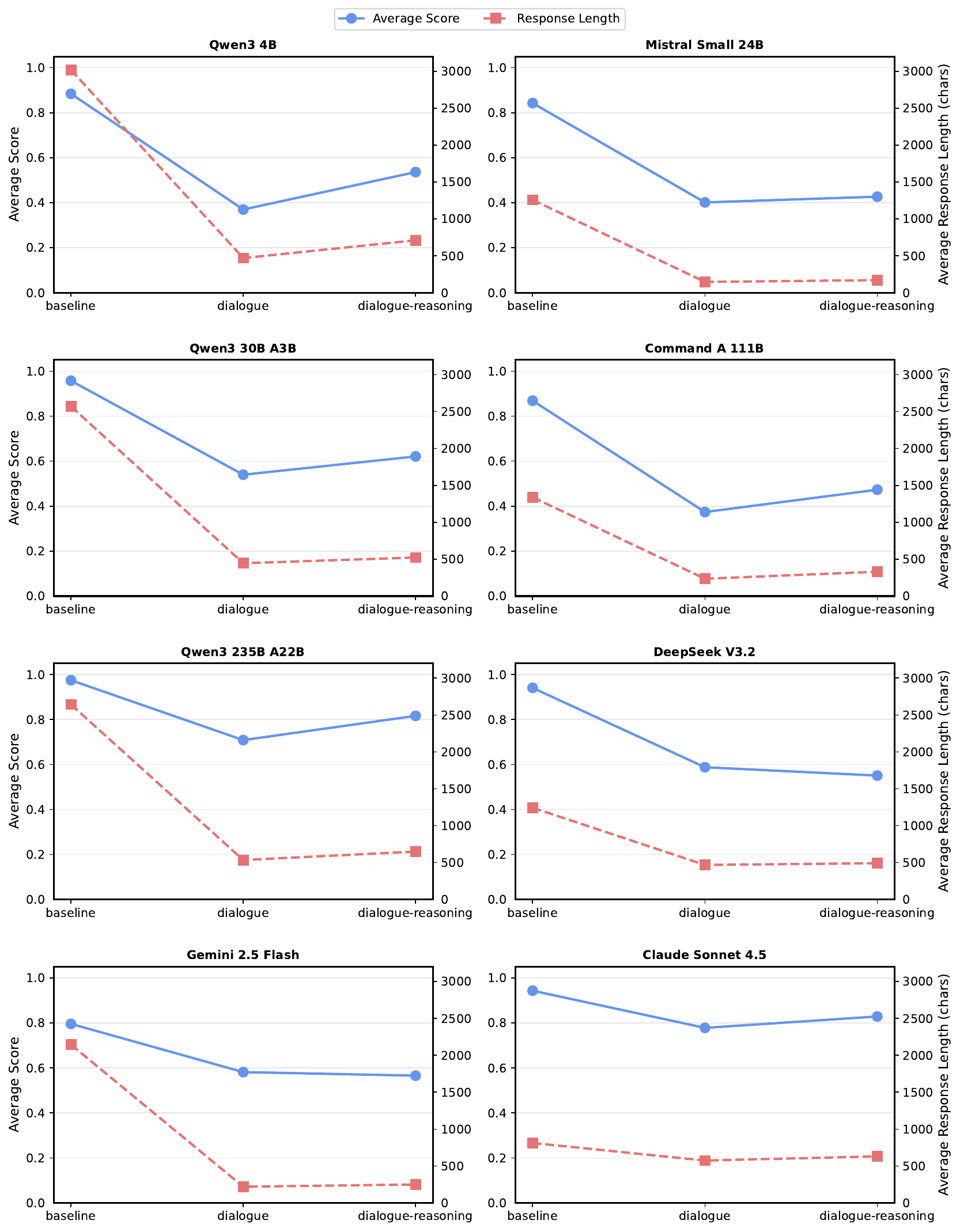}
    \caption{Comparison of scores and average response lengths in characters by LLM for the \textit{Dialogue with reasoning instructions} ablation.}
    \label{fig:accuracy_vs_length_reasoning}
\end{figure*}

\newpage
\begin{figure*}[h]
\centering
\small
\begin{tcolorbox}[
    colback=white,
    colframe=bordercolor,
    arc=3mm,
    boxrule=0.75pt,
    width=0.95\textwidth,
    left=5pt,
    right=5pt,
    top=5pt,
    bottom=5pt,
]
\ttfamily
\color{textgray}
\setlength{\parindent}{0pt}
\setlength{\parfillskip}{0pt plus 1fil}
\emergencystretch=3em
\sloppy

\textnormal{\textbf{Query}}
\vspace{0.5em}

Given the hotel information in JSON format below, calculate the total price for booking Limehouse for two nights for six people. Two people will each stay in their own single room and the others will share double rooms. One person in a double room will check in one day later.

[correct answer: £460]

\rule{\linewidth\color{bordercolor}}{0.4pt}

\vspace{1.0em}
\textnormal{\textbf{Baseline}}
\vspace{0.5em}

We are asked to calculate the **total cost for booking Limehouse for two nights for six people**,
with the following conditions:

- **Two people** will each stay in their **own single room**.

- The **remaining four people** will share **double rooms** (so two double rooms).

- **One person in a double room will check in one day later** (i.e., they only stay for **one
night**, not two).

\vspace{1.0em}
\#\#\# Step 1: Identify the hotel and prices

\vspace{0.5em}
The hotel is **Limehouse** (ID: 25), with the following pricing:

- **Single room**: £40 per night

- **Double room**: £75 per night

\vspace{1.0em}
\#\#\# Step 2: Break down the booking

\vspace{0.5em}
\#\#\#\# 1. Two single rooms (for two people staying for **2 nights** each)

- Cost per single room per night: £40

- For 2 nights: 40 × 2 = £80 per room

- For 2 rooms: 80 × 2 = **£160**

\vspace{1.0em}
\#\#\#\# 2. Two double rooms (for four people, but one person only stays one night)

\vspace{0.5em}
Let’s clarify the double room occupancy:

- Each double room holds 2 people.

- We have 4 people to accommodate in double rooms.

- But **one person checks in one day later**, so they only stay **1 night**.

So, we need to consider **how the rooms are used**:

- **Two double rooms** are booked for **2 nights** (since the hotel is booked for 2 nights).

- However, **one person** only stays for **1 night**, so we can assume they are in one of the double rooms for only one night.

But the **room is still booked for 2 nights**, so we must pay for the **full 2 nights** for each double room, **regardless** of how many people are in it each night.

So, we **cannot** charge only one night for the room — the room is reserved for 2 nights.

Therefore, **each double room costs £75 per night × 2 nights = £150**

- Two double rooms: 150 × 2 = **£300**

> Note: Even though one person only stays one night, the room is still occupied for two nights, so we pay full price.

\vspace{1.0em}
\#\#\# Step 3: Total cost

\vspace{0.5em}
- Two single rooms (2 nights each): £160  
- Two double rooms (2 nights each): £300  

**Total = 160 + 300 = £460**

\vspace{1.0em}
\#\#\# Final Answer:

\vspace{0.5em}
**£460** is the total cost for booking Limehouse for two nights for six people under the given conditions.

\rule{\linewidth\color{bordercolor}}{0.4pt}

\vspace{0.5em}
\textnormal{\textbf{Dialogue}}

\vspace{0.5em}
The total cost for your stay at Limehouse is £330. This includes:

- Two single rooms for two people (2 nights): £40 × 2 = £80

- Two double rooms for four people (2 nights): £75 × 2 = £150

- One double room for one person (1 night, starting on the 19th): £75

Total: £80 + £150 + £75 = £305

Wait, let me recalculate based on the booking confirmation. The system confirms the total is £330, which includes all the specified stays and room types.

So, the final total cost is **£330**. Let me know if you'd like to proceed with the booking!

\end{tcolorbox}
\caption{Comparison of responses from Qwen3 30B A3B for the \emph{Hotel booking price} task in the \emph{Baseline} and \emph{Dialogue} settings. In the \emph{Baseline} setup, the LLM provides correct answer, while it is incorrect in the \emph{Dialogue} setup. This is an example of the \emph{Post-hoc rationalization} pattern (see Section~\ref{sec:posthoc_rationalization}).}
\label{fig:example_early_answer}
\end{figure*}

\newpage
\begin{figure*}[h]
\centering
\small
\begin{tcolorbox}[
    colback=white,
    colframe=bordercolor,
    arc=3mm,
    boxrule=0.75pt,
    width=0.95\textwidth,
    left=5pt,
    right=5pt,
    top=5pt,
    bottom=5pt,
]
\ttfamily
\color{textgray}
\setlength{\parindent}{0pt}
\setlength{\parfillskip}{0pt plus 1fil}
\emergencystretch=3em
\sloppy

\textnormal{\textbf{Query}}
\vspace{0.5em}

How often on average do trains run between 20:00 and midnight?

[correct answer: every 19 minutes]

\rule{\linewidth\color{bordercolor}}{0.4pt}

\vspace{1.0em}
\textnormal{\textbf{Baseline}}
\vspace{0.5em}

To calculate how often trains run between 20:00 and midnight, we need to count the number of trains that depart within this time frame and then determine the average frequency.

\vspace{0.5em}
First, let's identify the trains that depart between 20:00 and midnight (23:59). We'll look at the `leaveAt` field in each train object.

\vspace{0.5em}
Here are the relevant trains:

1. TR13056: 20:07

2. TR13057: 20:21

3. TR13058: 20:37

4. TR13059: 21:07

5. TR13060: 21:21

6. TR13061: 21:37

7. TR13062: 22:07

8. TR13063: 22:21

9. TR13064: 22:37

10. TR13065: 23:07

11. TR13066: 23:21

12. TR13067: 23:37

\vspace{0.5em}
There are 12 trains departing between 20:00 and midnight.

\vspace{0.5em}
The time interval between 20:00 and midnight is 4 hours, which is equivalent to 240 minutes.

\vspace{0.5em}
To calculate the average frequency, we'll divide the total time interval (240 minutes) by the number of trains (12):

\vspace{0.5em}
Average frequency = Total time interval / Number of trains

= 240 minutes / 12

= 20 minutes

\vspace{0.5em}
So, on average, trains run every 20 minutes between 20:00 and midnight.

\vspace{0.5em}
**Answer:** Trains run approximately every **20 minutes** on average between 20:00 and midnight.

\rule{\linewidth\color{bordercolor}}{0.4pt}

\vspace{0.5em}
\textnormal{\textbf{Dialogue}}

\vspace{0.5em}

I'm sorry, I don't have access to the information required to answer your question. Is there anything else I can help with?

\end{tcolorbox}
\caption{Comparison of responses from Command A 111B for the \emph{Train departure frequency} task in the \emph{Baseline} and \emph{Dialogue} settings. In the \emph{Baseline} setup, the LLM provides an answer withing a minute of the target, while it refuses to solve the task in the \emph{Dialogue} setup. The response of an example of the \emph{Task avoidance} pattern (see Section~\ref{sec:task_avoidance}).}
\label{fig:example_refusal}
\end{figure*}

\newpage
\begin{figure*}[h]
\centering
\small
\begin{tcolorbox}[
    colback=white,
    colframe=bordercolor,
    arc=3mm,
    boxrule=0.75pt,
    width=0.95\textwidth,
    left=5pt,
    right=5pt,
    top=5pt,
    bottom=5pt,
]
\ttfamily
\color{textgray}
\setlength{\parindent}{0pt}
\setlength{\parfillskip}{0pt plus 1fil}
\emergencystretch=3em
\sloppy

\textnormal{\textbf{Query}}
\vspace{0.5em}

Is Cambridge Book And Print Gallery north of La Margherita?

[correct answer: yes]

\rule{\linewidth\color{bordercolor}}{0.4pt}

\vspace{1.0em}
\textnormal{\textbf{Baseline}}
\vspace{0.5em}

To determine whether **Cambridge Book And Print Gallery** is north of **La Margherita**, we need to compare their **y-coordinates** (northing) in the given spatial coordinate system, where the origin (0, 0) is in the **southwest** corner of the map.

\vspace{0.5em}
- In this system, **higher y-values** mean **further north**.

\vspace{1.0em}
\#\#\# Step 1: Get the coordinates

\vspace{0.5em}
From the JSON data:

\vspace{0.5em}
- **Cambridge Book And Print Gallery**  
  `location`: [22100, 8790] → **y = 8790**

- **La Margherita**  
  `location`: [22256, 10065] → **y = 10065**

\vspace{1.0em}
\#\#\# Step 2: Compare y-coordinates

\vspace{0.5em}
- 8790 (Cambridge Book And Print Gallery)

- 10065 (La Margherita)

\vspace{0.5em}
Since **8790 < 10065**, Cambridge Book And Print Gallery has a **lower y-coordinate**, meaning it is **south** of La Margherita.

\vspace{1.0em}

\#\#\# Final Answer:

\vspace{0.5em}

**No, Cambridge Book And Print Gallery is not north of La Margherita. It is south of it.**

\rule{\linewidth\color{bordercolor}}{0.4pt}

\vspace{0.5em}
\textnormal{\textbf{Dialogue}}

\vspace{0.5em}
Yes, Cambridge Book And Print Gallery is located north of La Margherita. While both are in the west area of Cambridge, the gallery is situated at a higher latitude (location [22100, 8790]) compared to La Margherita (location [22256, 10065]), meaning it's further north.

\end{tcolorbox}
\caption{Comparison of responses from Qwen3 235B A22B for the \emph{Directional relations} task in the \emph{Baseline} and \emph{Dialogue} settings. In the \emph{Baseline} setup, the LLM provides correct answer, while it is incorrect in the \emph{Dialogue} setup due an inversion of the coordinate system mapping provided in the prompt. The response is an example of the \emph{Rule inversion} pattern (see Section~\ref{sec:rule_inversion}).}
\label{fig:example_spatial}
\end{figure*}

\newpage
\begin{figure*}[h]
\centering
\small
\begin{tcolorbox}[
    colback=white,
    colframe=bordercolor,
    arc=3mm,
    boxrule=0.75pt,
    width=0.95\textwidth,
    left=5pt,
    right=5pt,
    top=5pt,
    bottom=5pt,
]
\ttfamily
\color{textgray}
\setlength{\parindent}{0pt}
\setlength{\parfillskip}{0pt plus 1fil}
\emergencystretch=3em
\sloppy

\textnormal{\textbf{Query}}
\vspace{0.5em}

What is the departure time of the latest train that arrives in Cambridge before sunset.

[correct answer: 14:09]

\rule{\linewidth\color{bordercolor}}{0.4pt}

\vspace{1.0em}
\textnormal{\textbf{Baseline}}
\vspace{0.5em}

I need to find the latest train that arrives in Cambridge before sunset (16:07).

\vspace{0.5em}
Let me check the arrival times:

- TR9225: arrives 12:54 (before 16:07)

- TR8882: arrives 13:54 (before 16:07)

- TR9758: arrives 14:54 (before 16:07)

- TR2361: arrives 15:54 (before 16:07)

- TR1575: arrives 16:54 (after 16:07)

- TR5424: arrives 17:54 (after 16:07)

- All subsequent trains arrive even later

\vspace{0.5em}
The latest train that arrives before sunset is **TR2361**, which arrives at 15:54.

\vspace{0.5em}
**The departure time is 14:09.**

\rule{\linewidth\color{bordercolor}}{0.4pt}

\vspace{0.5em}
\textnormal{\textbf{Dialogue}}
\vspace{1.0em}

Looking at the trains, the last departure that will get you to Cambridge before sunset (16:07 today) is:

\vspace{0.5em}
**Train TR9225**

- Departs Leicester: 11:09

- Arrives Cambridge: 12:54

- Duration: 105 minutes

- Price: £35.30 standard class / £52.90 first class

\vspace{0.5em}
This is actually the only train that arrives before sunset at 16:07. The next train (TR8882) arrives at 13:54, which is after sunset.

\vspace{0.5em}
Would you like me to help you purchase tickets for the 11:09 train?

\vspace{0.5em}

\end{tcolorbox}
\caption{Comparison of responses from Claude 4.5 Sonnet for the \emph{Train departure time} task in the \emph{Baseline} and \emph{Dialogue} settings. In the \emph{Baseline} setup, the LLM provides the optimal option, while it is suboptimal in the \emph{Dialogue} setup. The response is an example of the \emph{Boundary failure} pattern (see Section~\ref{sec:boundary_failures}).}
\label{fig:example_temporal}
\end{figure*}

\end{document}